%% file: main.tex
\documentclass[journal]{IEEEtai}

\usepackage[colorlinks,urlcolor=blue,linkcolor=blue,citecolor=blue]{hyperref}

\usepackage{color,array}

\usepackage{graphicx}
\usepackage{graphics}
\usepackage{amsmath}
\usepackage{wrapfig}
\usepackage[ruled,vlined]{algorithm2e}
\let\labelindent\relax
\usepackage{enumitem}

\usepackage{url}

\usepackage{multirow}

\DeclareMathOperator*{\argmax}{argmax}

\usepackage{amsthm}

\newtheorem{definition}{Definition}
\begin{document}

\input{sections/arxiv.tex}

\title{Gaussian Switch Sampling: A Second Order Approach to Active Learning}

\author{Ryan Benkert, \IEEEmembership{Student Member, IEEE}, Mohit Prabhushankar, \IEEEmembership{Member, IEEE}, Ghassan AlRegib, \IEEEmembership{Fellow, IEEE}, Armin Pacharmi, \IEEEmembership{Ford Motor Company}, and Enrique Corona, \IEEEmembership{Ford Motor Company} 
\thanks{This paragraph of the first footnote will contain the date on which you submitted your paper for review. It will also contain support information, including sponsor and financial support acknowledgment. For example, ``This work was supported in part by the U.S. Department of Commerce under Grant BS123456.'' }
\thanks{This paragraph will include the Associate Editor who handled your paper.}}

\markboth{Subitted to Journal of IEEE Transactions on Artificial Intelligence}
{Ryan Benkert \MakeLowercase{\textit{et al.}}: Gaussian Switch Sampling: A Second Order Approach to Active Learning}

\maketitle

\begin{abstract}
In active learning, acquisition functions define informativeness directly on the representation position within the model manifold. However, for most machine learning models (in particular neural networks) this representation is not fixed due to the training pool fluctuations in between active learning rounds. Therefore, several popular strategies are sensitive to experiment parameters (e.g. architecture) and do not consider model robustness to out-of-distribution settings. To alleviate this issue, we propose a grounded second-order definition of information content and sample importance within the context of active learning. Specifically, we define importance by how often a neural network "forgets" a sample during training - artifacts of second order representation shifts. We show that our definition produces highly accurate importance scores even when the model representations are constrained by the lack of training data. Motivated by our analysis, we develop Gaussian Switch Sampling (\texttt{GauSS}). We show that \texttt{GauSS} is setup agnostic and robust to anomalous distributions with exhaustive experiments on three in-distribution benchmarks, three out-of-distribution benchmarks, and three different architectures. We report an improvement of up to 5\% when compared against four popular query strategies. Our code is available at \href{https://github.com/olivesgatech/gauss}{https://github.com/olivesgatech/gauss}.
\end{abstract}

\begin{IEEEImpStatement}
With the ever increasing demand for deep learning products in safety-critical applications, the acquisition of suitable training data has significantly increased in complexity. In several instances, labeling large quantities of data is associated with insurmountable costs (e.g. medical applications) while other instances require data diversity at scale with numerous edge cases (e.g. autonomous vehicles). Active learning offers a promising solution to both of these problems by selecting data to both improve annotation efficiency and data quality. For practical deployment, these algorithms must select robust datasets and further function in a wide variety of training setups in order to guarantee design requirements. However, existing algorithms base their acquisition function on the representation approaches which results in high performance fluctuations over training setups and robustness metrics. For instance, an algorithm might perform exceedingly well with one neural network architecture but underperform with another. Our work introduces a second-order active learning approach for robustness and portability. We see our work as the first step of many in bringing theoretical active learning algorithms to real world deployment.
\end{IEEEImpStatement}

\begin{IEEEkeywords}
Active Learning, Learning Dynamics, Example Forgetting
\end{IEEEkeywords}

\input{sections/Introduction}
\input{sections/RelatedWork}
\input{sections/Background}
\input{sections/Methodology}
\input{sections/Results}
\input{sections/Conclusion}

\section*{Acknowledgments}
The authors would like to thank the remaining members of OLIVES and the anonymous reviewers for their feedback. In particular, we would like to thank Dr. Gukyeong Kwon for fruitful discussions within the preliminary phase of the project. This work was funded by a Ford-Gerogia Tech Alliance Project.

\bibliographystyle{plain}
\bibliography{mybib}

\begin{IEEEbiography}[{\includegraphics[width=1in,height=1in,clip,keepaspectratio]{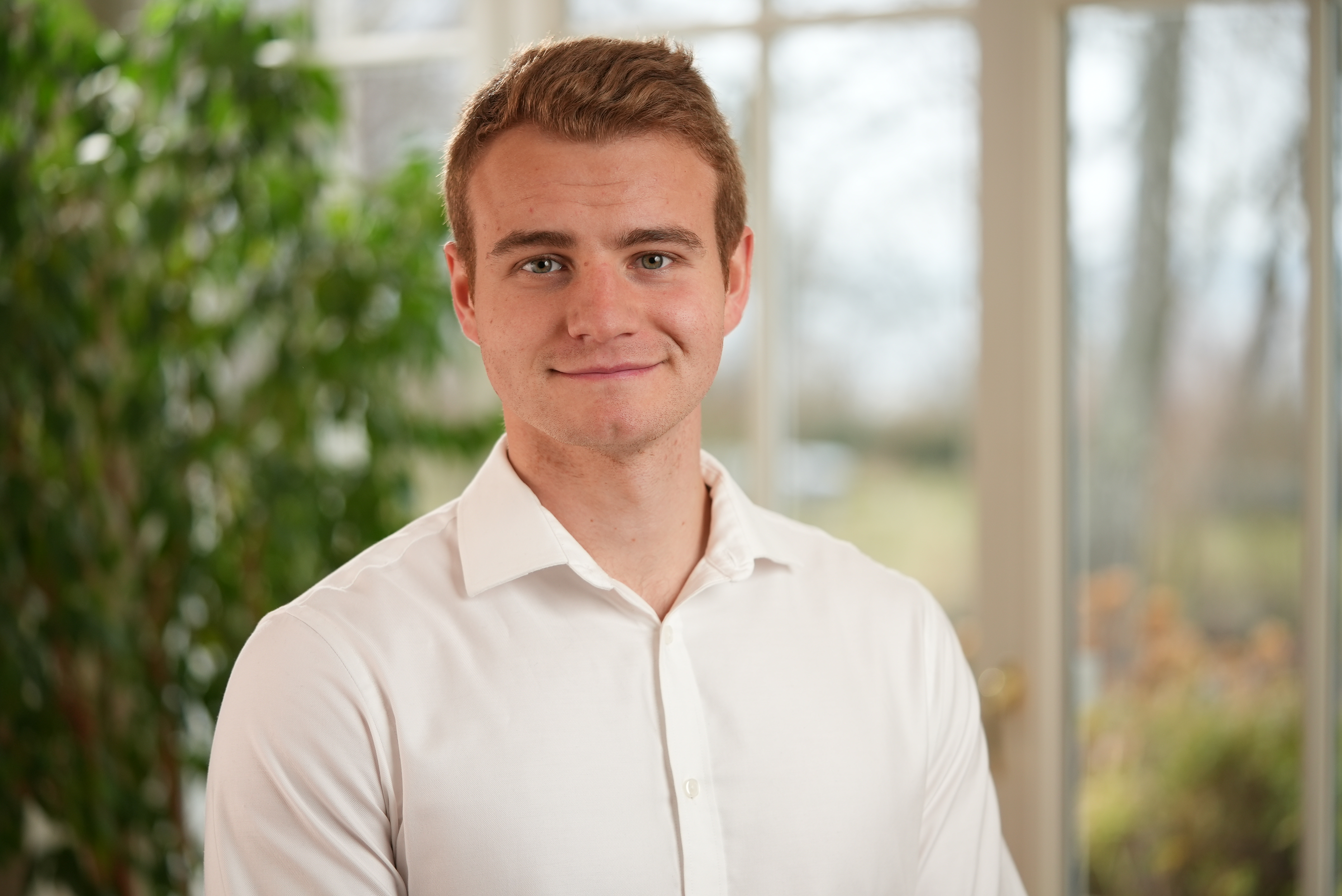}}]{Ryan Benkert}{\space} is a Ph.D. student in the Omni Lab for Intelligent Visual Engineering and Science (OLIVES) at the Georgia Institute of Technology. In his research, he addresses fundamental challenges in machine learning that bridge the gap between academic research and industrial deployment. His interests include active learning, uncertainty estimation, and neural network learning dynamics. Prior to Georgia Tech, he received his B.Sc and M.Sc from the RWTH Aachen University in Germany. In his free time, Ryan dances west coast swing and is an avid sports enthusiast. 
\end{IEEEbiography}

\begin{IEEEbiography}[{\includegraphics[width=1in,height=1in,clip,keepaspectratio]{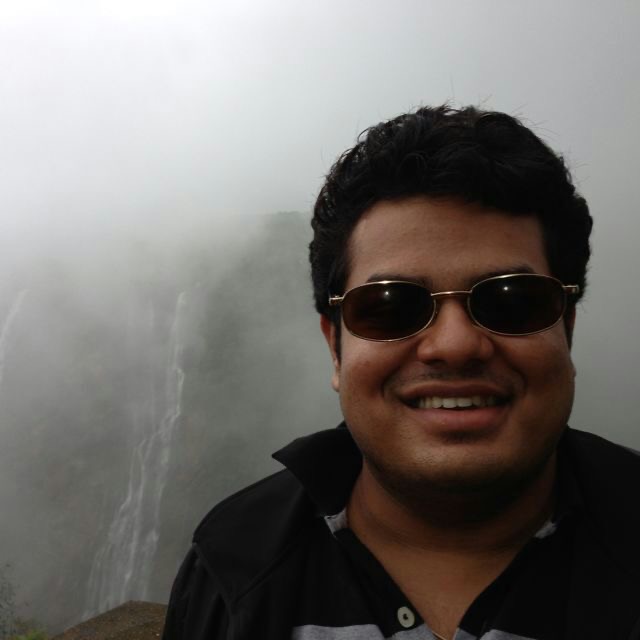}}]{Mohit Prabhushankar}{\space} received his
Ph.D. degree in electrical engineering from the Georgia
Institute of Technology (Georgia Tech), Atlanta, Georgia,
30332, USA, in 2021. He is currently a Postdoctoral Fellow in the School of Electrical and Computer Engineering at the Georgia Institute of Technology in the Omni Lab for Intelligent Visual Engineering and Science (OLIVES) lab, working in the fields of image processing, machine learning, and explainable and robust AI. He is the recipient of the Best Paper award at ICIP 2019 and Top Viewed Special Session Paper Award at ICIP 2020. He is the winner of the Roger P Webb ECE Graduate Research Excellence award in 2022. He has served as a Teaching Fellow at Georgia Tech since 2020. He is an IEEE Member.
\end{IEEEbiography}

\begin{IEEEbiography}[{\includegraphics[width=1in,height=1in,clip,keepaspectratio]{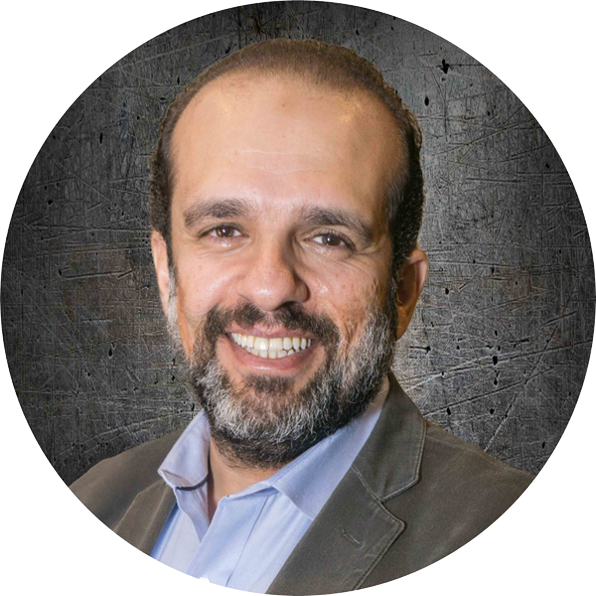}}]{Ghassan AlRegib}{\space} received his Ph.D. in electrical engineering from the Georgia Institute of Technology (Georgia Tech), Atlanta, Georgia, 30332, USA. He is currently the John and Marilu McCarty Chair Professor in the School of Electrical and Computer Engineering at the Georgia Institute of Technology. He was a recipient of the ECE Outstanding Graduate Teaching Award in 2001 and both the CSIP Research and the CSIP Service Awards in 2003, the ECE Outstanding Junior Faculty Member Award, in 2008, and the 2017 Denning Faculty Award for Global Engagement. His research group, the Omni Lab for Intelligent Visual Engineering and Science (OLIVES) works on research projects related to explainable machine learning, robustness in intelligent systems, interpretation of subsurface volumes, and expanding healthcare access and quality. He has participated in several service activities within the IEEE and served on the editorial boards of several journal publications. He served as the TP co-Chair for ICIP 2020 and GlobalSIP 2014. He served as expert witness on several patents infringement cases and advised several corporations on both technical and educational matters. He is IEEE Fellow.  
\end{IEEEbiography}

\begin{IEEEbiography}[{\includegraphics[width=1in,height=1in,clip,keepaspectratio]{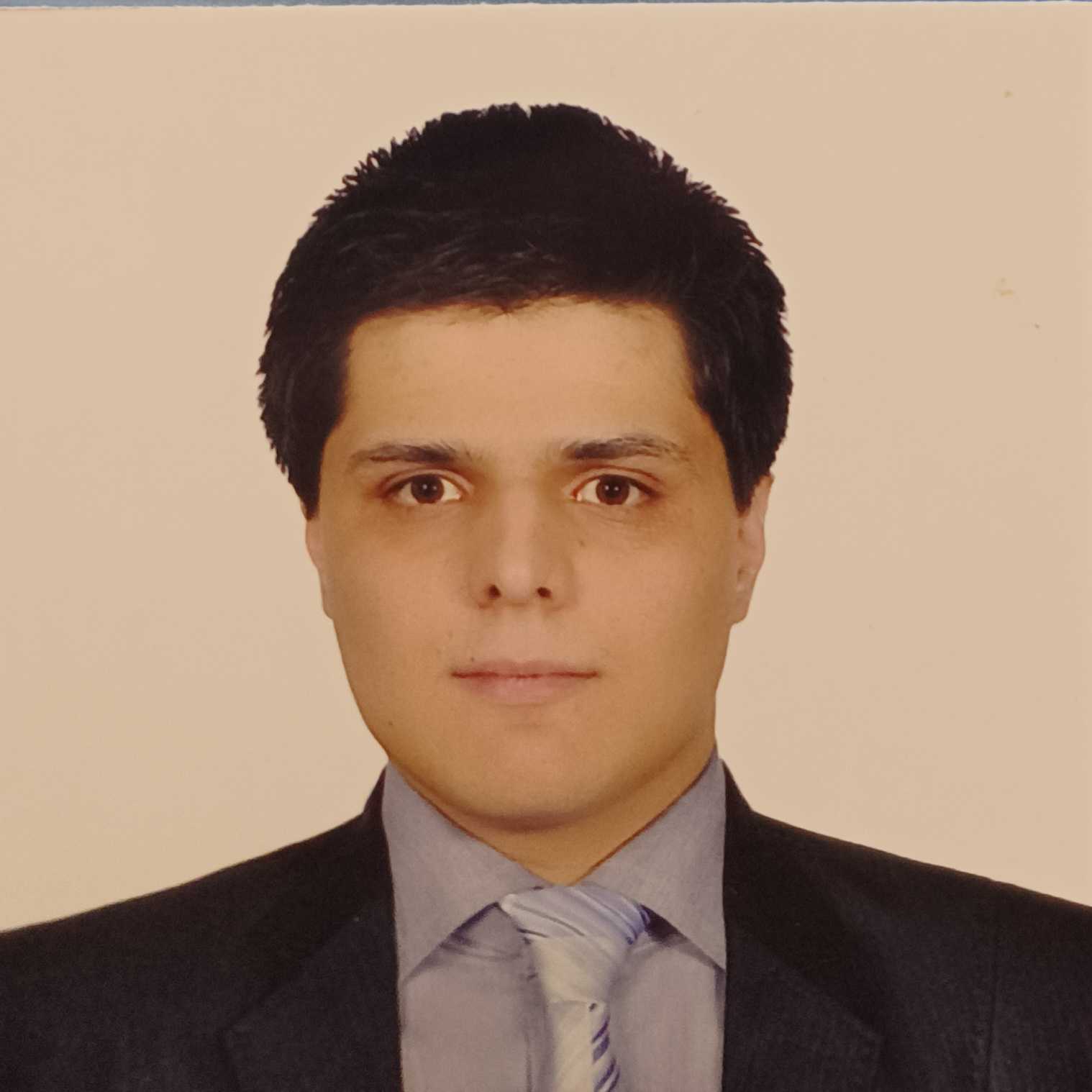}}]{Armin Parchami}{\space} recieved his B.E. in Software Engineering and M.Sc. in Artificial Intelligence from Bu-Ali Sina University and then he received his Ph.D. in Computer Science from UTA in 2017. His dissertation was on single shot face recognition using deep learning algorithms for security applications. Between 2017 and 2022, he was working at Ford on level 4 autonomous vehicles. He is currently managing the perception team at Ford ADAS developing perception algorithms for L2+ autonomy. His current research interests include monocular 3D object detection, active learning, and wide baseline sensor fusion.
\end{IEEEbiography}

\begin{IEEEbiography}[{\includegraphics[width=1in,height=1in,clip,keepaspectratio]{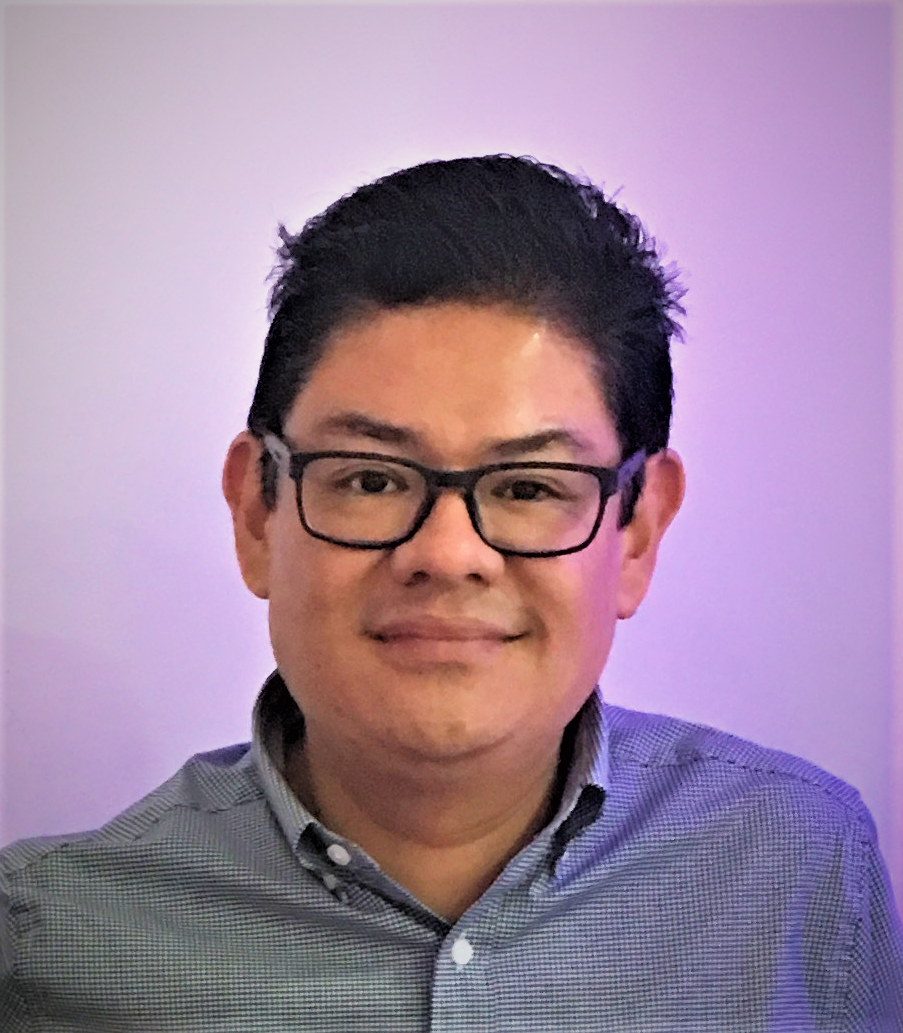}}]{Enrique Corona}{\space} received his B.S.E.E. from Universidad de las Americas Puebla, Mexico with a joint M.Sc. in Electrical and Computer Engineering. He was awarded a Ph.D. in Electrical Engineering from Texas Tech University in 2012. His dissertation was about unsupervised learning methods for applications to segmentation of medical images using kernel mappings to improve clustering model selection. He then worked in the design and implementation of intelligent home appliances using computer vision until 2016. After joining Ford Motor Co. he started working in the development of technology for smart city infrastructure and L4 autonomous vehicles. At present he is part of Ford ADAS collaborating in L2+ autonomy projects. He is currently interested in monocular 3D object detection and tracking, multiple view and geometric computer vision, and kernel methods. 
\end{IEEEbiography}

\end{document}

%% file: sections/arxiv.tex
\onecolumn 

\begin{description}[labelindent=-1cm,leftmargin=1cm,style=multiline]

\item[\textbf{Citation}]{R. Benkert, M. Prabhushankar, G. AlRegib, A. Parchami, and E. Corona, “Gaussian Switch Sampling: A Second Order Approach to Active Learning,” in IEEE Transactions on Artificial Intelligence (TAI), 2023} \\


\item[\textbf{Review}]
{
Date of submission: 4 June 2022\\
Date of first revision: 4 November 2022\\
Date of second revision: 8 January 2023\\
Date of Acceptance: 5 February 2023
} \\


\item[\textbf{Bib}] {@ARTICLE\{benkert2022\_TAI,\\ 
author=\{R. Benkert, M. Prabhushankar, G. AlRegib, A. Parchami, and E. Corona\},\\ 
journal=\{IEEE Transactions on Artificial Intelligence\},\\ 
title=\{Gaussian Switch Sampling: A Second Order Approach to Active Learning\}, \\ 
year=\{2023\}\\ 
} \\


\item[\textbf{Copyright}]{\textcopyright 2022 IEEE. Personal use of this material is permitted. Permission from IEEE must be obtained for all other uses, in any current or future media, including reprinting/republishing this material for advertising or promotional purposes,
creating new collective works, for resale or redistribution to servers or lists, or reuse of any copyrighted component
of this work in other works. }
\\
\item[\textbf{Contact}]{\href{mailto:rbenkert3@gatech.edu}{rbenkert3@gatech.edu}  OR \href{mailto:alregib@gatech.edu}{alregib@gatech.edu}\\ \url{http://ghassanalregib.info/} \\ }
\end{description}

\thispagestyle{empty}
\newpage
\clearpage
\setcounter{page}{1}

\twocolumn

%% file: sections/Introduction.tex
\section{Introduction}
A core factor in the success of machine learning algorithms is the selection of suitable data. While several samples contain valuable information for a given task, other samples may be redundant with little additional information or anomalous with contradicting features. Active Learning~\cite{cohn1996active, settles2009active} is a paradigm in machine learning that selects the most informative samples from a large unlabelled data pool for annotation and training. Due to its intuitive practicality, active learning has already impacted multiple industrial sectors including manufacturing \cite{tong2001active}, robotics \cite{alrobotics}, recommender systems \cite{alrecommender}, medical imaging \cite{hoi2006batch}, and autonomous vehicles \cite{haussmann2020scalable}.

At the core of every active learning approach stands its method to select the next set of annotation samples - its acquisition function. The function establishes a ranking based on \emph{information content} and selects interactively from the most informative samples. Therefore, the success of an acquisition function heavily relies on 1) the definition of high information content and on 2) the information content approximation when the model representations are constrained due to the lack of annotated training data. In other words, a successful acquisition function relies on \emph{what} it considers ``informative" and \emph{how} well the model produces importance scores. Hence, an effective definition is both accurate in defining sample importance and simple to approximate by a constrained model.

\begin{figure}[!h]
    \begin{center}
        \includegraphics[scale=0.37]{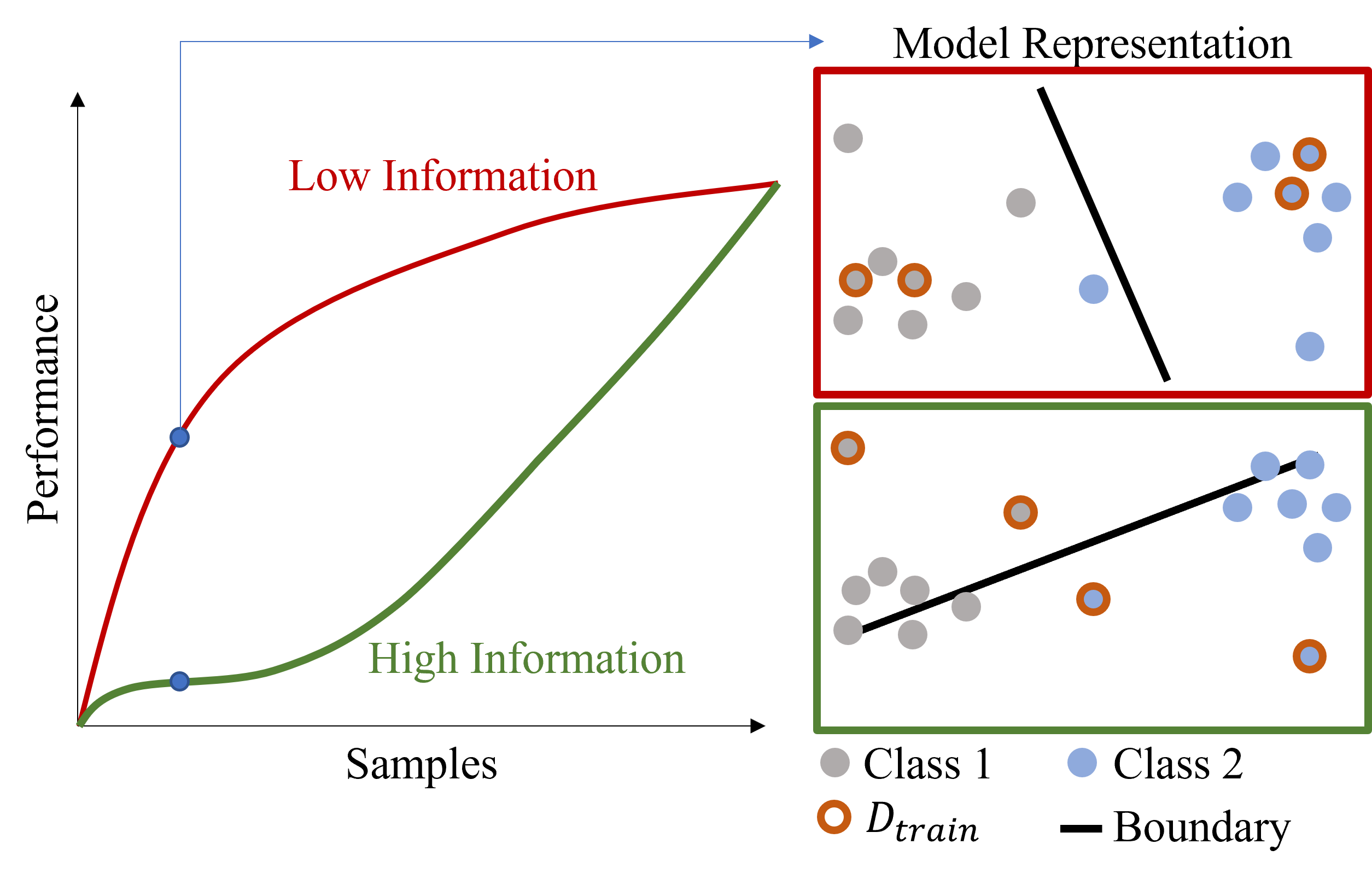}
    \end{center}
    \caption{Toy example of active learning with samples containing high/low information content. With a high information content outliers are selected for the training set $D_{train}$ and can result in inaccurate decision boundaries.}
    \label{fig:information-toy}
\end{figure}

Intuitively, we would assume that an effective definition of informativeness would result in superior generalization performance. In our analysis, we find this assumption to be incorrect. In fact, definitions that produce accurate importance estimates can even hurt generalization performance for early active learning rounds when the model is heavily constrained. For a simple explanation, consider a toy classifier that distinguishes cats from dogs. The samples with the highest information content are frequently anomalous or subject to data noise. In our example, this could be an image where both a cat and a dog are present or the presence of a rare breed that significantly differs from the remaining data points. Within the context of active learning, training on highly informative (or anomalous) samples results in an inaccurate representation space and thus a lower generalization performance on the test set. We sketch such a scenario in Figure~\ref{fig:information-toy} and further show this behavior empirically in later sections. With acquisition functions that produce ``too accurate" information content scores, the resulting model overfits to outlier data and therefore results in lower performance. Even though several existing approaches do not consider this essential design argument, they can still be effective in practical scenarios. In particular, existing algorithms (willingly or unwillingly) introduce noise within the acquisition function that results in inaccurate information content estimates and hence better performance. However, existing approaches define sample importance on the model representation directly which is noisy in early rounds due to the lack of data. As an example, entropy sampling \cite{wang2014new} selects samples with the highest softmax entropy, a direct manifestation of the model representation. 

\begin{figure}[!h]
    \begin{center}
        \includegraphics[scale=0.37]{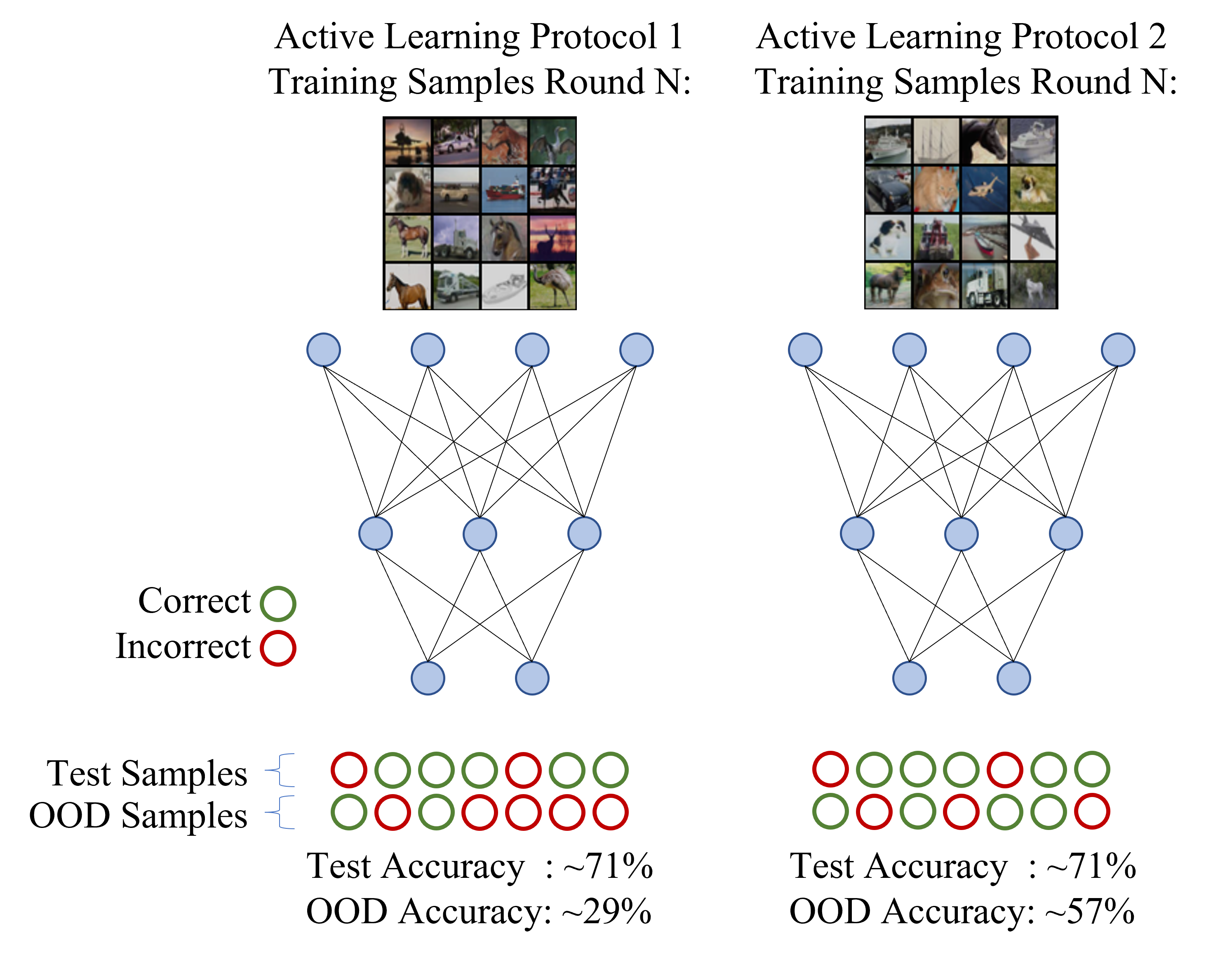}
    \end{center}
    \caption{Toy example of out-of-distribution performance in active learning. Both protocols result in similar test performance but severely differ when deployed on out-of-distribution samples.}
    \label{fig:ood-toy}
\end{figure}

Even though existing methods can be effective, relying on the representation directly has two hazardous implications: First, the acquisition function performance is highly dependent on experimental parameters (e.g. architecture) that directly influence the model representation. For this reason, several methods perform well for one parameter constellation but not in another. Second, the acquisition function does not consider model robustness to samples that do not originate from the training distribution (out-of-distribution). In particular, the model is incapable of reliably selecting informative samples that increase robustness due to the inherently unreliable model representations of outliers \cite{hendrycks2019robustness}. 
We show a toy example of two different protocols exposed to out-of-distribution settings in Figure~\ref{fig:ood-toy}. In active learning round~N, we show two different protocols with similar test set performance. However, when exposed to out-of-distribution samples, the performance of both protocols significantly diverges and protocol~2 outperforms protocol~1 by a large margin. While papers on active learning generalization are ubiquitous, out-of-distribution active learning studies are significantly less common \cite{kothawade2021similar, benkert2022forgetful}. For this purpose, popular active learning protocols lack robustness qualities and performance is unpredictable, especially in out-of-distribution scenarios. In fact, our experiments in Section~\ref{sec:nemerical evaluation} show significant performance variations of popular active learning approaches in out-of-distribution settings.

\begin{figure*}[!h]
    \begin{center}
        \includegraphics[scale=0.50]{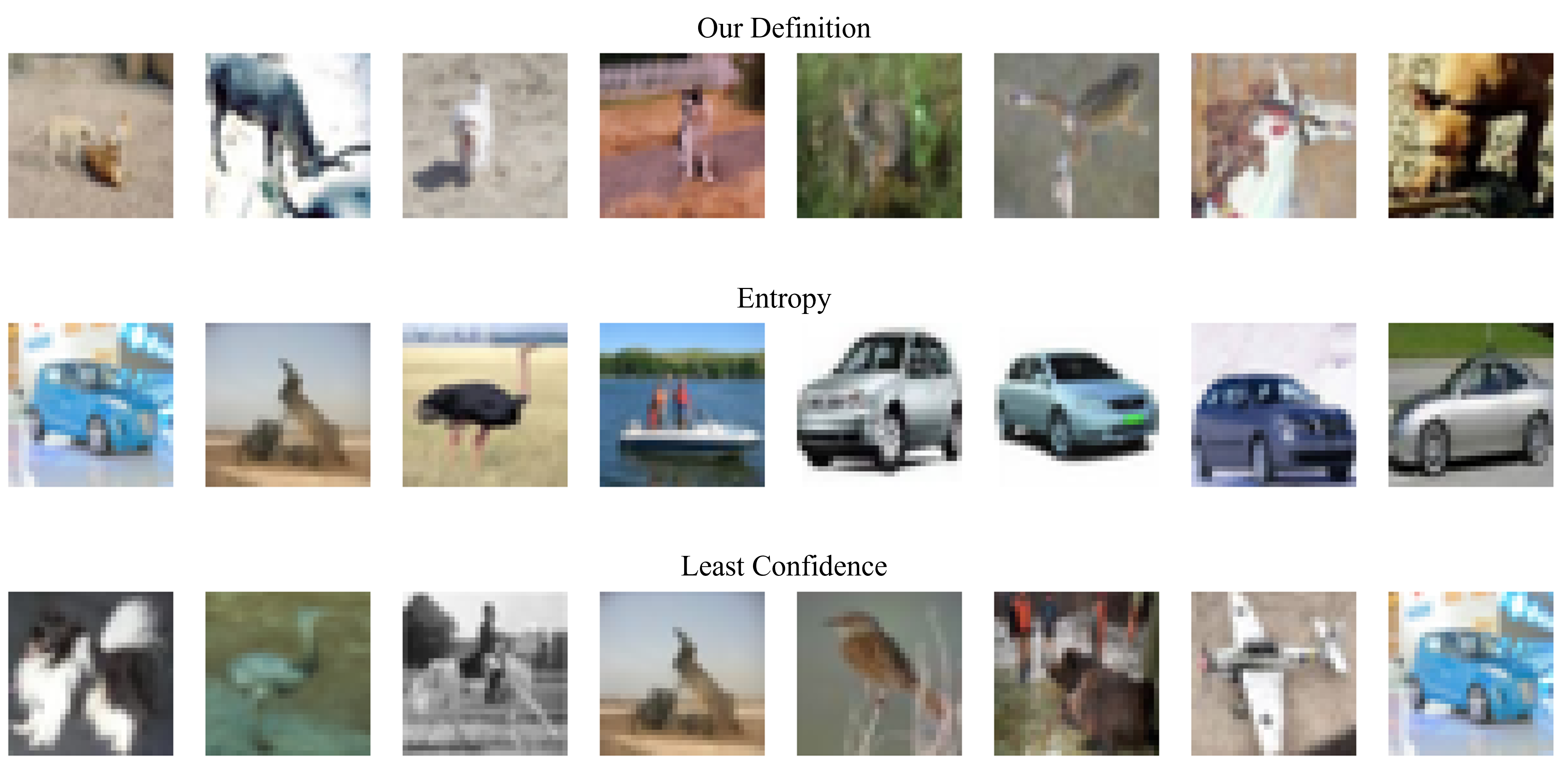}
    \end{center}
    \caption{Examples of informative samples when using different definitions of information content. The examples are generated without limited data constraints. From top to bottom: Our definition, entropy sampling, least confidence sampling.}
    \label{fig:examples}
\end{figure*}

In this paper, we propose a grounded approach to defining information content in active learning. Specifically, we consider a second-order approach where ``informative" is defined by representation shifts rather than the representation directly. We consider a sample most informative if it was ``forgotten" the most during training. In practice, evaluating ``forgetting" requires annotations which are not present within the unlabeled pool. For this purpose, we approximate ``forgetting" as a switch within the sample prediction. This characteristic is especially important for selecting samples that make our model more robust to out-of-distribution samples. In a detailed analysis, we show that our practical ``switch" definition is both exceedingly accurate in assessing sample importance as well as simple to approximate by a constrained model. To showcase the quality of our approach, we show samples with a high information content in Figure~\ref{fig:examples} according to different definitions of information content. Specifically, we show highly informative samples as by our definition and two popular definitions in active learning literature. We see that our informative samples are exceedingly harder to distinguish or are ambiguous which implies a higher information content. For instance, our ``informative" samples are ambiguous images with complex shapes or ambiguous colors (e.g. the frog image in the top row; third image from the right side). In contrast, the other definitions consider clear images informative that are highly similar among each other. For instance, entropy contains five similar car images that are clearly distinguishable. 

In order to provide favorable generalization performance, we further extend our method with an interactive sampling technique based on a gaussian mixture model. Our approach produces a targeted noise vector that successfully induces enough noise for model generalization while biasing the selection batch to highly informative samples that improve model robustness. We call our method \texttt{Gaussian Switch Sampling} or \texttt{GauSS} in short. We validate our algorithm with exhaustive experiments and compare against popular protocols used in practical active learning pipelines. Overall, we find that \texttt{GauSS} is robust to setup changes, performs favorably in out-of-distribution settings, and achieves up to 5\% improvement in terms of accuracy over popular existing strategies.

%% file: sections/RelatedWork.tex
\section{Related Work}
\subsection{First Order and Second Order Active Learning}
In active learning, a strong research branch involves exploring different acquisition functions and their definitions of information content. In this context, the majority of existing approaches leverage the model representation directly to define the next acquisition batch. Due to the direct dependency, we refer to these approaches as first-order acquisition functions. To the best of our knowledge, second-order approaches remain largely unexplored.

Overall, active learning approaches differ in their definition of information content for samples within the unlabeled data pool. In this context, several approaches define information content with generalization difficulty. As an example, several approaches define sample importance using softmax probabilities \cite{wang2014new, roth2006margin} where information content is related to the output logits of the network. For instance, entropy sampling queries samples that have the highest softmax entropy while least confidence sampling queries samples with the lowest prediction probability (or confidence) score. Closely related to this concept, the authors in~\cite{schohn2000less, tong2001support} query samples based on decision boundary proximity. For instance, \cite{tong2001active} query samples with support vector machines. \cite{gal2017deep} define importance as model uncertainty and select the following sample batch based on Monte Carlo dropout. Finally \cite{Beluch_2018_CVPR} use an ensemble of classifiers to query the next set of samples. In contrast, other approaches define information content based on data representation within the dataset. Acquisition functions in this field are based on the assumption that representative samples best approximate the overall dataset structure and improve performance. In this context, a large group of approaches focus on constructing the core-set of the unlabeled data pool \cite{sener2017active, longtailcoreset}. Furthermore, \cite{gissin2019discriminative} use a discriminative approach, where the authors reformulate the active learning problem as an adversarial training problem. Specifically, they formulate active learning as a binary classification problem and select samples that minimize the differences between the labeled and unlabeled pool. Finally, there are several approaches that consider the combination of both data representation and generalization difficulty within their definition of information content. In several cases \cite{ash2019deep, haussmann2020scalable}, an ordering is established based on generalization difficulty and the representative component is introduced by sampling interactively from the ordering. \cite{batchbald} integrate both generalization difficulty and data representation by extending \cite{houlsby2011bayesian} to diverse batch acquisitions. Finally, \cite{baram2004online, hsu2015active} deploy "mix and match" meta-active learning approaches, that switch strategies each active learning round.

Even though several approaches are effective, they directly depend on the model representation of the data and are susceptible to setup shifts or out-of-distribution samples. 

\subsection{Out-of-Distribution Analysis}
Within the field of deep learning, the importance of out-of-distribution samples within the context of robustness is well established \cite{hendrycks2019robustness}. In this context, several approaches involve measurements of uncertainty \cite{gal2015bayesian, kendall2017uncertainties, van2020uncertainty}, gradient representations \cite{kwon2019distorted, kwon2020backpropagated, kwon2020novelty, lee2021open, gradientsuncertainty, prabhushankar2020contrastive, prabhushankar2021contrastive, alregib2022explanatory}, or model calibration \cite{guo2017calibration, prabhushankar2021introspective}. For instance, \cite{van2020uncertainty} reject out-of-distribution samples by ranking samples uncertainty scores from a deterministic neural network. \cite{kwon2020backpropagated}, detect anomalies with by imposing a gradient constraint on the model that distinguishes inliers from outliers, \cite{guo2017calibration} introduce temperature scaling as an effective method to calibrate neural networks. With a few exceptions, out-of-distibution scenarios remain largely unexplored within the context of active learning with the scarce examples \cite{kothawade2021similar} being first-order approaches.

\subsection{Efficient Active Learning and Theory}
Apart from developing new active learning acquisition functions, there has been extensive research regarding efficient active learning as well as its practical implementation. \cite{coleman2019selection} improve computational efficiency by approximating the target model with a smaller proxy model. Further, \cite{haussmann2020scalable} consider scalability issues of active learning for object detection. In particular they investigate the implications of varying diverse and uncertain active learning components on large-scale data.

A substantial research direction is further dedicated to theoretical aspects of active learning. For instance, \cite{wang2018optimal, ting2018optimal, han2016local} consider subsampling in regression problems. Several works also focus on understanding the active learning process itself. In this context, \cite{farquhar2021statistical} investigate the implicit bias inherent within the active learning process.

In this work, we explore the capabilities of different protocols to assess information content. This represents a substantial contribution for designing efficient active learning paradigms as well as theoretic explanations for a acquisition function performance.

\subsection{Continual Learning}
In addition to active learning, our work is related to the field of continual learning and catastrophic forgetting \cite{chen2018lifelong, kirkpatrick2016overcoming, ritter2018online, toneva2018empirical}. In this context, significant research efforts involve reducing forgetting through model constraints \cite{aljundi2018memory, kirkpatrick2016overcoming}, loss exploration \cite{shi2021overcoming}, or augmentation \cite{benkert2021explaining, benkert2022example}. While the approaches effectively reduce forgetting, they do not consider (or estimate) information content in active learning settings

Most significantly, our work shares a strong connection with the study of \cite{toneva2018empirical} on forgetting events. Within the article, the authors quantify neural networks forgetting in the training set and empirically show dependencies between forgetting and several important machine learning topics such as dataset compression and label noise. Even though the paper provides an interesting framework for quantifying neural network forgetting, the quantification requires labels and is unsuitable for information content estimation in practical active learning settings. In contrast, our work expands the concepts of \cite{toneva2018empirical} to function without annotation and develop a grounded definition of information content for active learning.

%% file: sections/Background.tex
\section{Defining Sample Information with Learning Dynamics}
Active learning consists of iteratively selecting a set of unlabeled samples for annotation. We refer to a single iteration as active learning round. In this section we formulate information content as learning difficulty within each active learning round. Intuitively, we define informative as the frequency in which a network ``forgets" unlabeled samples during training. We introduce active learning and neural network forgetting singularly and further analyze our definition in comparison to existing definitions of informativeness.
\subsection{Active Learning}
In active learning, the objective is to improve data efficiency iteratively by involving the model in the data annotation process. In this context, consider a dataset $D$ where a small subset $D_{train}$ represents the initial training data and $D_{pool}$ represents the unlabelled data pool for the selection process. The goal within each active learning round is to select a batch of $b$ samples  $X^* = \{x^*_1, ..., x^*_b\}$ that improves the model accuracy the most when added to the training data $D_{train}$. The selection of $X^*$ can be defined as

\begin{equation}
\label{eq:aquisition-function}
    X^* = \argmax_{x_1, ..., x_b \in D_{pool}} a(x_1, ..., x_b | f_w(x))
\end{equation}

where $a$ represents the acquisition function and $f_w$ is the fully trained deep model conditioned on the parameters $w$. In this context, we define first-order algorithms and second-order algorithms by the manner in which they are conditioned on $f_w$:

\begin{definition}[First- and Second-Order Active Learning]
We define an active learning algorithm as first-order when the acquisition function is conditioned on the fully trained neural network $f_w$ exclusively. Further, an algorithm is of second-order when the acquisition function is conditioned on dynamic fluctuations of $f_w$ caused by external interference (e.g. training).
\end{definition}

\subsection{Defining Importance with Forgetting Events}
\label{sec:learning-dynamics-in-active-learning}
In this paper, we define importance through the learning difficulty of different unlabeled samples. Specifically, we build upon \cite{toneva2018empirical} and measure information content with the frequency in which samples are "forgotten" during training. Intuitively, a sample is "forgotten" if it was classified correctly ("learnt") at time $t$ and subsequently misclassified ("forgotten") at a later time $t' > t$.\\ 

Without loss of generalizability, we consider recognition tasks where the objective is to predict a label $\tilde{y_i}$ of sample $x_i$ that corresponds to the ground truth annotation $y_i$. The accuracy of the model for sample $x_i$ at an arbitrary time $t$ can be defined as,

\begin{equation}
\label{eq:accuracy}
    acc^{t}_{i} = \mathbf{1}_{\tilde{y}^t_i = y_i}.
\end{equation}

In Equation~\ref{eq:accuracy}, $\mathbf{1}_{\tilde{y}^t_i = y_i}$ describes a binary variable that indicates whether the classification was correct at time $t$. Specifically, it reduces to one if $x_i$ was correctly predicted or zero if the sample was misclassified. Based on our notation, we define a sample as "forgotten" if the accuracy decreases within two subsequent time steps:

\begin{equation}
f_{i}^{t} = int( acc^{t}_{i} < acc^{t-1}_{i} ) \in [{1, 0}].
\end{equation}

Here, $f_i^t$ is a binary result of accuracy variables and reduces to one if the accuracy within subsequent optimization steps decreases (a switch from correctly to incorrectly classified). Similar to \cite{toneva2018empirical}, we define the binary event $f_i^t$ as a \emph{forgetting event} at time $t$. Based on these terminologies, we define the information content of each sample in a simple and intutive manner as follows:

\begin{definition}[Information in Active Learning]
Within the context of active learning, we quantify information by the amount of forgetting events that occur for a given sample each round. Specifically, we consider samples with the largest amount of forgetting events as the most informative while fewer forgetting events imply redundant information.
\end{definition}

In contrast to existing definitions, forgetting events are not based on the representation directly but are second-order artifacts of decision boundary shifts. We reason that this characteristic is favorable for distribution shifts and setup choices.


\subsection{Prediction Switches}
Even though forgetting events are simple to formulate and provide an intuitive measure for importance, the computation is not tractable for practical active learning protocols. Specifically Equation~\ref{eq:switch-events} requires the label annotation $y_i$ which is unavailable for the unlabeled set $D_{pool}$ by definition. For this purpose, we approximate forgetting events with prediction switches. A prediction switch occurs when the model output prediction $\tilde{y}$ changes within time intervals. Formally, we write
\begin{equation}
\label{eq:switch-events}
s^{t}_{i} = \mathbf{1}_{\tilde{y}^t_i != \tilde{y}^{t-1}_i}.
\end{equation}

Analogous to our previous definition, we call $s^t_i$ a \emph{switch event} at time $t$.

\section{Information Definitions in Active Learning}
In this section, we investigate the extent in which prediction switches differ to existing importance definitions. Specifically, we want to answer 1) how effectively do prediction switches estimate information content? and 2) how accurate are the importance scores when the representation is constrained?

In practical settings, sample representations are constrained by the size of the training set $D_{train}$ and can result in inaccurate importance quantification, especially when the definition is based on the representation directly. For this purpose, we conduct our analysis by artificially removing the constraint imposed by limited data and base our importance scores on fully trained model representations - i.e. the output of the acquisition function is based on representations derived from a model trained on both $D_{train}$ and $D_{pool}$. Even though this analysis puts our method at a clear disadvantage we note that our importance rankings are significantly more accurate than existing protocols. 

\subsection{Accurate Importance Scores vs. Performance Trade-off}
In active learning, acquisition functions define sample importance by the information content added to the dataset. However, informative samples in practice are frequently irregular and can result in insufficient data representations when trained on informative samples exclusively. Using our cat/dog example classifier from the introduction, a training set with informative samples could exclusively contain rare cat/dog breeds that are underrepresented within the underlying distribution. A classifier trained on this constrained dataset will result in low generalization performance on a test set containing mostly well represented breeds. Within the context of active learning, acquisition functions that render importance scores that are ``too accurate" frequently result in significantly \emph{lower} performance than the random baseline or other acquisition functions with ``less accurate" importance scores. 

In our analysis, we consider scenarios where the representation space is not constrained by limited data and therefore significantly improves the importance score accuracy of the compared strategies. For this reason, effective practical acquisition functions perform significantly lower than the random baseline and exhibit \emph{lower} generalization performance with \emph{higher} information content within the acquisition batch.



\subsection{Importance in Optimal Settings}
\label{sec:optimal-strat-lcs}
We investigate which acquisition functions provide the most informative samples when acquisition functions produce accurate measurements. For this purpose, we consider an optimal setting where we first train a model on the full CIFAR10 training data and derive the sample representations from the fully trained model. The following importance scores (e.g. softmax entropy) are subsequently calculated with the optimal sample representations. For our proposed definition, we count the switch events while training on the full training data and query the samples ordered from most to least switch events. In this section, we compare optimal settings only as the discussion on practical implementations with limited annotation access is thoroughly discussed in Section~\ref{sec:experiments}.

\begin{figure*} [ht!]
    \begin{center}
        \includegraphics[scale=0.45]{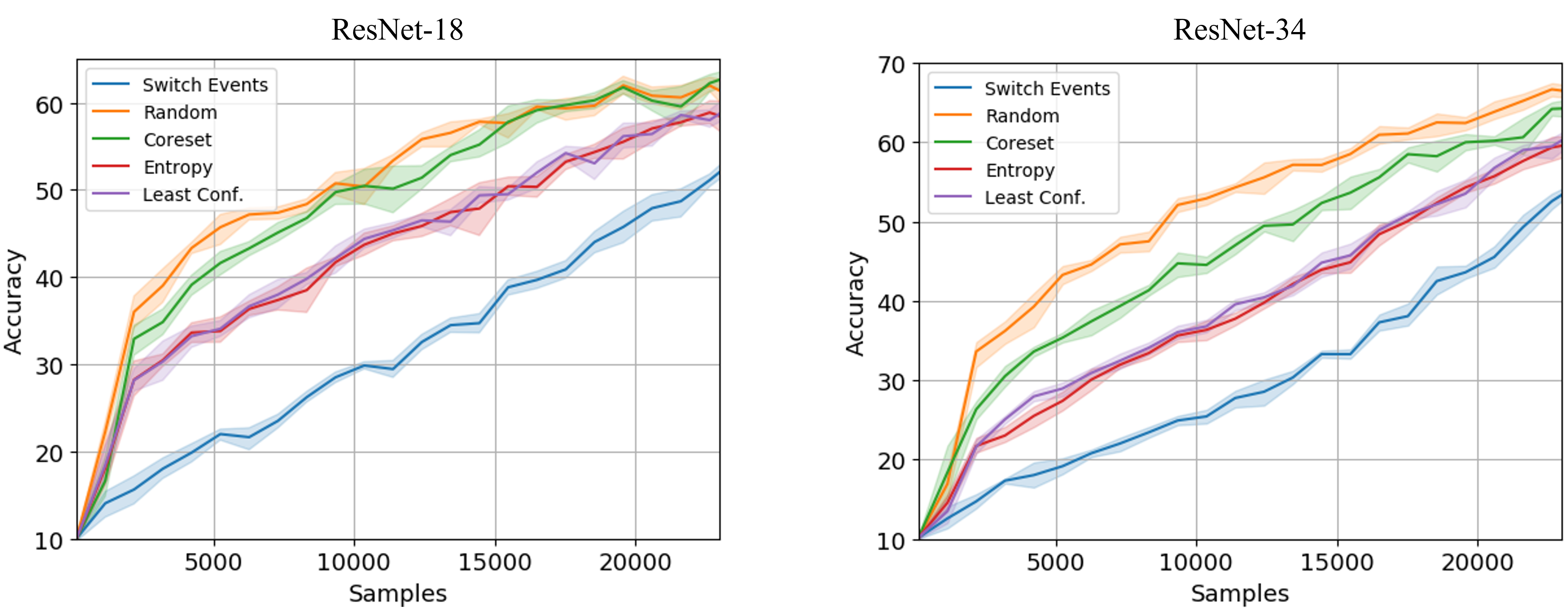}
    \end{center}
    \caption{Different query strategies in with optimal sample representations. We compare random sampling, entropy sampling, least confidence sampling, coreset, as well as switch events.}
    \label{fig:optimal-strats}
\end{figure*}

In all of our active learning experiments, we start with an initial training set $D_{train}$ of 128. In each round, we query 1024 additional samples with the repsective active learning strategy. For our active learning models we use a resnet-18, as well as a resnet-34 architecture \cite{he2016deep}. We optimize with the adam variant of SGD and a learning rate of $10^{-4}$. No augmentations are used for our active learning experiments. 
For our optimal representations, we train a network on the full CIFAR-10 dataset for 200 epochs and extract the representations for our active learning experiments. For the fully trained model, we optimize with SGD and use a multi step learning rate scheduler for training. Specifically, our initial learning rate is 0.1 and we divide the learning rate by 5 in epochs 100, 125, 150, and 175 respectively. 
We further use the augmentations random crop, random horizontal flip, and cutout. We choose this setup to derive competitive representations for the acquisition functions. We show the results of our analysis in Figure~\ref{fig:optimal-strats}. Specifically, we compare switch events to entropy sampling \cite{wang2014new}, random sampling, least confidence sampling \cite{wang2014new}, and coreset \cite{sener2017active}. We note that querying with optimal representations results in significantly lower generalization performance than the random baseline due to the high importance score accuracy. This is especially true for early rounds where only a limited amount of training data is available for the active learning model. Samples with a high information content are typically aberrant outliers that are disruptive to the representation space. Therefore, a more significant deviation from random sampling implies a higher information content. We note, that querying with switch events results in the most profound difference compared to the other strategies and implies the highest information content within each query. It is important to note that most existing active learning protocols outperform random sampling in practical settings - i.e. when the sample representations are inaccurate due to limited training data. In these cases, limited representation capabilities ``help" the acquisition function by introducing a random noise factor within the query. Several strategies even introduce targeted sampling techniques to counteract accurate information scores \cite{ash2019deep, haussmann2020scalable}. In GauSS we do this by sampling from a gaussian mixture model. 

\subsection{Approximation of Optimal Representations}
\begin{figure}[!h]
    \begin{center}
        \includegraphics[scale=0.35]{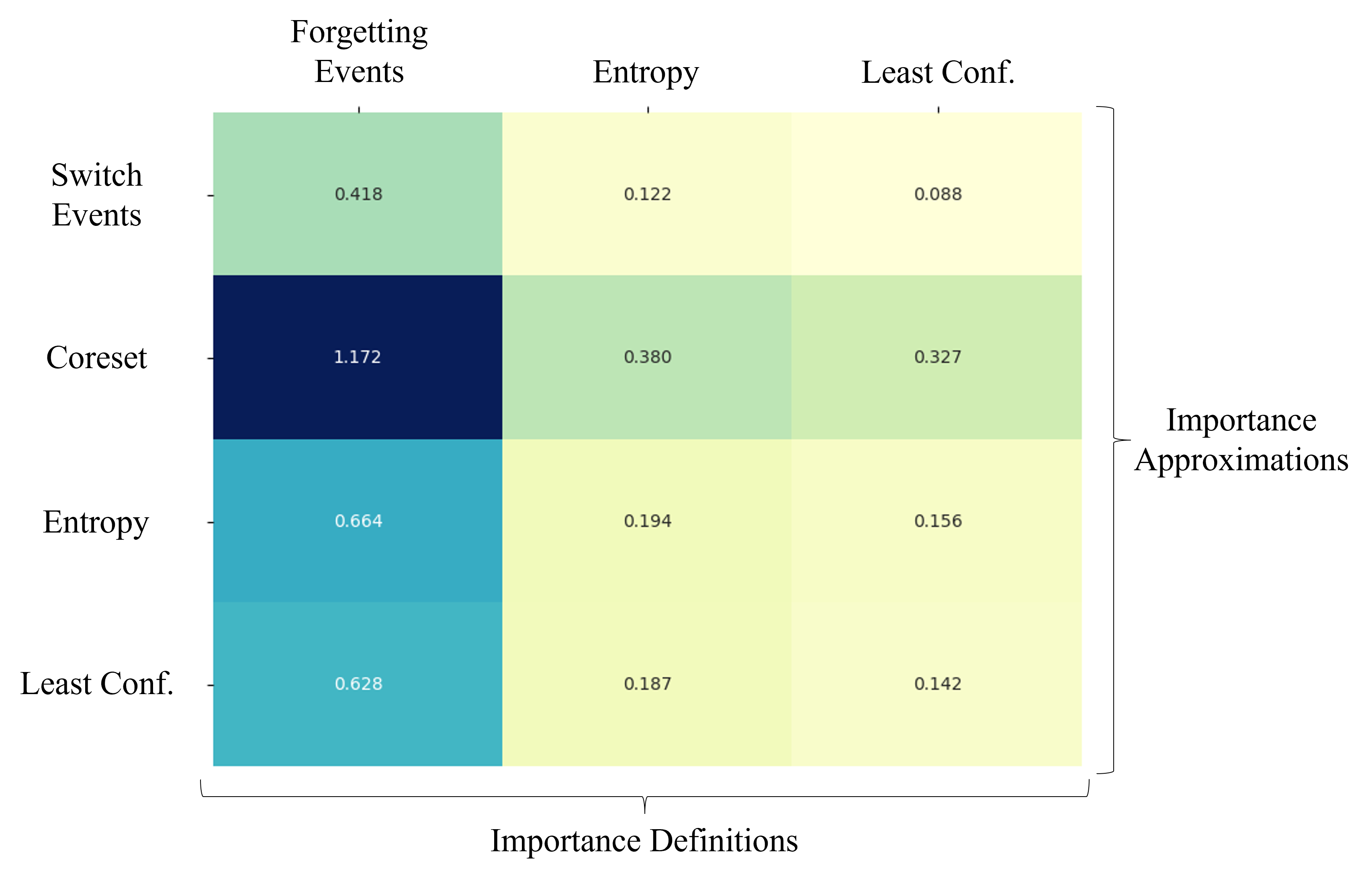}
    \end{center}
    \caption{Average KL-Divergence between the forgetting events distributions over 20 active learning rounds. We compare query strategies with their capability to approximate optimal settings with different importance definition. Columns represent optimal settings and rows show their practical approximations. Low scores indicate a high similarity in protocol behavior.}
    \label{fig:kl-dist}
\end{figure}

We further investigate the approximation quality of different importance definitions with respect to their optimal counter part. As described within the previous section, most existing strategies outperform the random baseline while their optimal counter part significantly underperforms in terms of generalization accuracy. We follow that accuracy alone is not a sufficient metric to capture the optimal approximation capabilities of an active learning protocol. For this purpose, we measure protocol similarity through information content within the training set of each active learning round. In this context, we evaluate the importance score for each sample within the training set and compare the distributions of the importance scores for each round through statistical difference metrics. If the samples contain similar information, the resulting importance scores render similar distributions and vice versa. Specifically, we perform the following steps for each active learning round: 1) We evaluate the importance score for every training sample. 2) We estimate the distribution by creating histograms over the importance scores for every acquisition function. 3) We evaluate the statistical difference between different acquisition function pairs (in our case we use a smoothed Kullback-Leibler divergence). 
We show our results in Figure~\ref{fig:kl-dist}. The different rows show the active learning protocols while the columns show the optimal counter parts. We compare switch events against entropy sampling and least confidence sampling as these are the most competitive definitions in our previous analysis. Further, we use forgetting events as the optimal counterpart of our switch event protocol. We choose this setup, as switch events should not only approximate their optimal switch event counterpart but should ideally approximate our original importance definition based on sample forgetting, not prediction switches. In our experiments, we use the same active learning setup as the previous section for experiments with optimal representations; The non-optimal protocol setup is identical with the exception of the data representation source - i.e. we follow the normal protocol workflow and gather sample representations from the model trained on $D_{train}$. We further generate importance scores for the histograms with forgetting events and compare histograms by calculating the Kullback-Leibler divergence smoothed by a gaussian filter. We average our result over the first 20 rounds. 

\begin{figure*}[!ht]
    \begin{center}
        \includegraphics[scale=0.50]{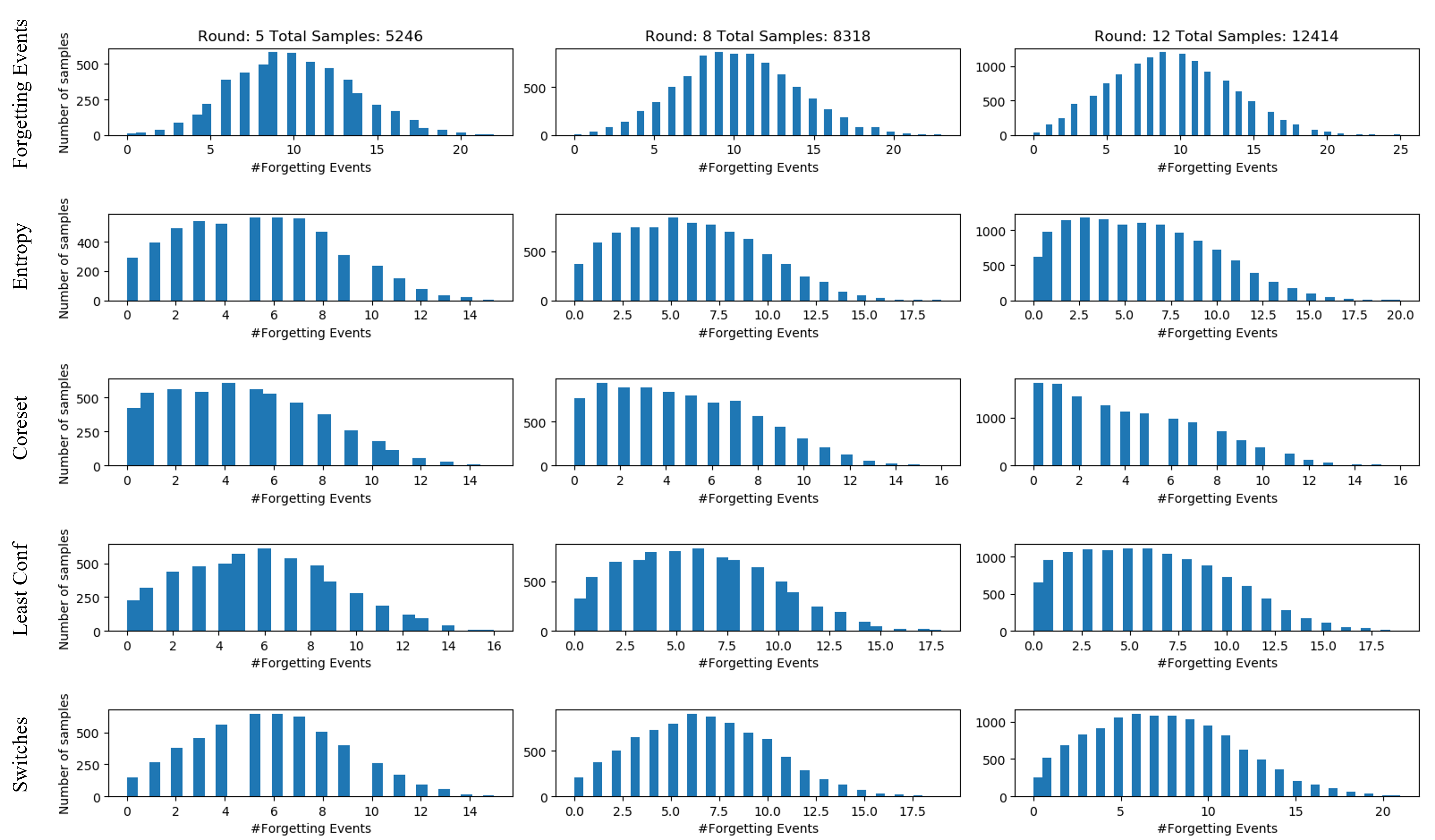}
    \end{center}
    \caption{Histograms for different definitions of information content in different rounds. The first row, samples data points with the highest forgetting events within an optimal setting - i.e. the importance scores are derived from a model trained on the fully labeled training set. The remaining rows are practical active learning scenarios where the importance scores are derived from a model with a constrained training pool. This includes entropy sampling, coreset, least confidence sampling and prediction switches.}
    \label{fig:nideal-hist}
\end{figure*}

Overall, we note that switch events show the closest distribution similarity to all optimal strategies indicating the most accurate information content approximation within the round-wise training sets. We reason that switch events are not dependent on the model representation directly and are therefore more accurate in identifying the most informative batch regardless of the importance definition. This further manifests itself visually in the distribution histograms of individual rounds (Figure~\ref{fig:nideal-hist}). We see that switch events exhibit the visually most similar distribution to optimal forgetting events distribution than the other strategies. In particular, we note that the other strategies query more samples with less amounts of forgetting events than the switch event acquisition function which results in different distribution shapes. As a final note, we add that even though we compare distributions with forgetting events we observe a similar behavior when comparing against other importance definitions.

%% file: sections/Methodology.tex
\section{Gaussian Switch Sampling}
Based on our initial analysis, we reason that switch events are effective for estimating information for unlabeled samples. Further, we reason that the second-order nature is potentially more robust to different training setups and out-of-distribution settings. However, we noted in our experiments that effective importance scores bias queries towards outliers that inhibit generalization capabilities. For this purpose, we force targeted random noise into our acquisition batch by probabilistic sampling with a bias towards higher switch events. Specifically, we model the switch event distribution with a two component gaussian mixture model and sample from the component with the higher switch event mean. 

Building on the classification of \cite{toneva2018empirical} into forgettable and unforgettable data, we model each sample within the unlabeled sampling pool to be either less-switching $U$ or frequently switching $F$. The total amount of switch events $S_{total}$ can then be modeled as the sum of the switch events from both less switching $S_U$ and frequently switching samples $S_F$:

\begin{equation}
\label{eq:sevents-rv-modeling}
S_{total} = S_U + S_F
\end{equation}

When we query from the unlabeled pool, we would ideally target the most informative samples or, more specifically, the samples originating from class $F$. 
For this reason, we fit a gaussian mixture model with two components (one for $U$ and one for $F$) to the switch event distribution and assign the less-switching distribution and frequently switching distribution to the gaussian mixture with the lower and higher mean respectively. During selection, we distinguish frequently switching samples from less switching samples with the separate gaussian components and sample the next acquisition batch from the frequenly switching distribution. Mathematically, we model the switch event distribution as

\begin{equation}
\label{eq:rv-modeling}
S_{total} \sim \omega_u N_u(\mu_u, \sigma_u) + \omega_f N_f(\mu_f, \sigma_f)
\end{equation}

where $\omega_{u/f}, \mu_{u/f}$, and $\sigma_{u/f}$ refer to the weight, mean, and  standard deviation of the singular distribution components $U$ and $F$ respectively. We subsequently establish the acquisition batch by sampling data points with probabilities from the second switch event distribution component $N_f$:

\begin{equation}
\label{eq:acq-rv-modeling}
X^* \sim N_f(\mu_f, \sigma_f)
\end{equation}

\begin{figure}[!h]
    \begin{center}
        \includegraphics[scale=0.50]{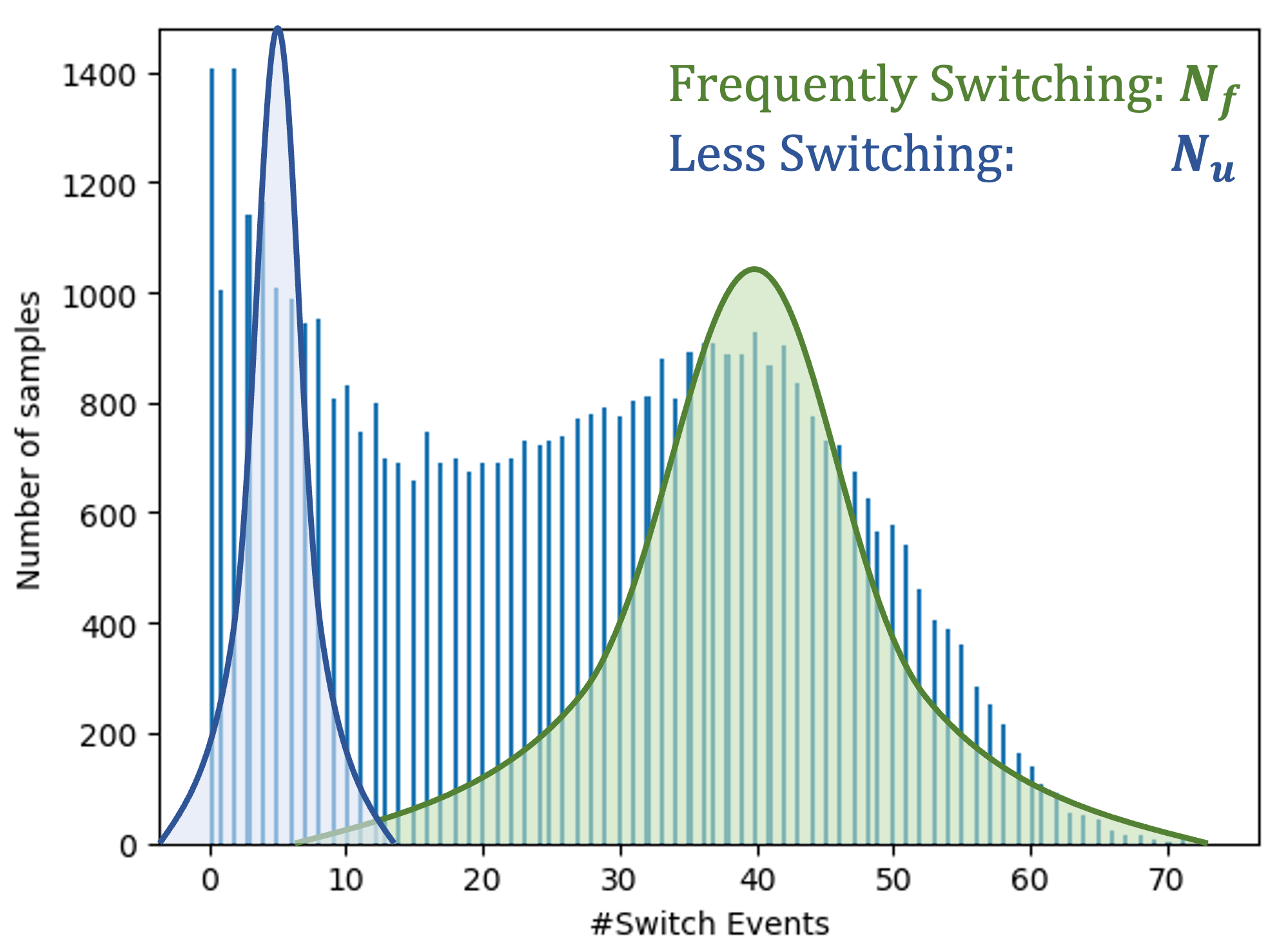}
    \end{center}
    \caption{Toy example of gaussian switch sampling. We sketch the switch event distribution within a single active learning round, as well as two gaussian mixture components. The next batch is generated by sampling from the most switching component in green.}
    \label{fig:rv-toy}
\end{figure}

In Figure~\ref{fig:rv-toy}, we show a toy example of our approach. Specifically, we show a histogram over the switch events occurring for a given round and sketch the ideally fitted gaussian mixture components over the histogram distribution. In this case, we would sample from the right (green) gaussian component. We call our method \emph{Gaussian Switch Sampling} or \texttt{GauSS} in short. 

%% file: sections/Results.tex
\section{Experiments}
\label{sec:experiments}
With our experiments we want to answer two central questions: 1) How well does \texttt{GauSS} perform in conventional active learning where the only representation constraint is imposed by $D_{train}$?, and 2) How does the active learning performance change in out-of-distribution scenarios? We feel that both 1) and 2) are realistic scenarios in practical deployment and represent a comprehensive study of the generalization capabilites of \texttt{GauSS}.

\subsection{Numerical Comparison}
\label{sec:nemerical evaluation}
In our experiments, we compare \texttt{GauSS} against four popular active learning protocols. Specifically, we compare against entropy sampling \cite{wang2014new}, coreset \cite{sener2017active}, active learning by learning (ALBL) \cite{hsu2015active}, least confidence sampling \cite{wang2014new}, and BatchBald \cite{batchbald}. We choose this constellation as it provides a large variety of importance definitions: Entropy sampling and least confidence sampling define sample importance with generalization difficulty while coreset maximizes the diversity of the training pool; active learning by learning represents a bandit style approach that interactively switches between coreset and least confidence during each round. It represents a fusion protocol that combines both generalization difficulty and data diversity. Finally, Batchbald is a recent bayesian approach that extends the popular BALD \cite{houlsby2011bayesian} method.  

We consider three out-of-distribution settings as well as three in-distribution settings. 
For our out-of-distribution analysis, we sample our training set from the CIFAR10 training set and measure accuracy on CIFAR10-C \cite{hendrycks2019robustness}, STL10 \cite{stl10}, and CINIC10 \cite{cinic10}. 
We choose these benchmarks, as they contain examples of data corruption and domain shifts - both of which are common in practical machine learning pipelines: Both STL10 and CINIC10 represent examples of cross domain out-of-distribution. STL10 is a difficult dataset with 8000 test samples and nine out of ten overlapping classes with CIFAR10 (we only consider the overlapping classes for accuracy calculations). CINIC10 is significantly larger than CIFAR10 with a test set of 90000 images and fully overlapping classes with CIFAR10. Furthermore, CIFAR10-C is an artificially corrupted version of CIFAR10 with 20 different corruption types on five different severity levels. We consider all corruptions except ``labels", ``shot noise", and ``speckle noise" and test on level two as well as level five each round. We choose this setup as it includes realistic data corruptions on a wide variety of severity levels. Our in-distribution analysis considers the datasets CIFAR10, CINIC10, and CURE-TSR \cite{Temel2017_NIPSW}. Both CIFAR10 and CINIC10 are popular in-distribution benchmarks with increasing difficulty and dataset size (60000 images for CIFAR10 and 270000 images for CINIC10). 
CURE-TSR represents the practically relevant application of traffic sign recognition with corrupted and uncorrupted training and test sets. The dataset is especially challenging as traffic signs occur at different frequencies resulting in a high class imbalance. Hence, an effective algorithm must perform on both well represented as well as underrepresented classes to improve generalization performance on the test set. In our experiments, ee consider the uncorrupted training and test set only. 
For all datasets except CURE-TSR, we start with an initial training pool of 128 randomly chosen samples and iteratively query 1024 samples each round according to the respective protocols. 
For BatchBald, we implement a bayesian model by applying monte-carlo droupout. Similar to the setup of \cite{batchbald}, we apply a dropout layer with a dropout probability of 0.4 before the final fully connected layer and sample 70 monte-carlo instances.
When experimenting with CURE-TSR, we start with an initial training pool of 32 samples and query an additional 32 samples each round. We choose this setup as it significantly increases the difficulty for traffic sign recognition. We further use three popular network architectures: Resnet-18 \cite{he2016deep}, resnet-34 \cite{he2016deep}, and densenet-121 \cite{densenet}. Similar to our previous experiments, we optimize with the Adam variant of SGD with a learning rate of $10^{-4}$ and no learning rate scheduler. No data augmentations are used and we use pre-trained model weights for our out-of-distribution experiments as well as our experiments on the CIFAR10 dataset. We reason that this represents a practical scenario where data is abundant in one domain but scarce in another. Finally, we retrain our model from scratch each round to prevent warm starting \cite{ash2020warm}. For all constellations, our results are averaged over 5 random seeds. Overall, this amounts to a total of 450 separate active learning experiments (we count all CIFAR10-C datasets as one experiment).

In several cases, the performance of active learning strategies differs across selection rounds. For instance, a strategy may show a strong accuracy increase for early rounds, and underperform in later rounds. Therefore, we quantify algorithm performance by summarizing the margin in which it outperforms (or underperforms) the random baseline. With random baseline, we refer to sampling random data points from the unlabeled pool each round.  First, we calculate the difference to the random baseline for each round by subtracting the random baseline accuracy curve from the accuracy curve of the respective strategy. Second, we combine the difference values by integrating over the subtraction curve across several active learning rounds. 
In short, the metric amounts to the difference in accuracy area compared to the random baseline. For reference, we provide the absolute area of the random baseline in the first row of each architecture.
A negative difference (random accuracy is higher) contributes negatively to the integral and positive differences (comparison accuracy is higher) contributes positively to the integral. Hence, a negative integral indicates an overall higher random accuracy curve while a positive integral indicates an overall outperformance with respect to the random baseline. In addition, the higher the integral the more significant the positive difference and the larger the outperformance margin. 
By quantifying performance with differences to the same reference point, we remove biases from absolute values in each round. For instance, the difference in early rounds contributes to the integral with the same magnitude as later rounds even though the absolute accuracy is higher in later stages.
For our results on CIFAR10-C, we average the difference curve accross all corruptions and levels. We show the integration results in Table~\ref{table:results-overall}.


We note, that \texttt{GauSS} shows the overall best performance over both in-distribution, and out-of-distribution benchmarks in nearly every experiment constellation. In particular, we note that \texttt{GauSS} shows an overall better performance than the random baseline in nearly every constellation. In contrast, the other strategies can outperform or underperform the random baseline. For instance, entropy sampling outperforms the random baseline in several cases (e.g. CIFAR10-C with resnet-34), but underperforms in other cases (e.g. CIFAR10 in-distribution test set with densenet-121). Moreover, we note that \texttt{GauSS} is more consistent accross different dataset and architecture choices. In all instances except a few outliers, \texttt{GauSS} outperforms the random baseline. In contrast, we find high fluctuations with competitive protocols. For instance, least confidence sampling significantly outperforms the random baseline on CIFAR10-C and CIFAR10 when using resnet-18, but underperforms when switching the architecture to densenet-121. We reason that \texttt{GauSS} queries more consistently as sample queries and model representation are not directly coupled. If the model representation severely suffers within a specific experiment configuration (e.g. when densenet-121 is used instead of resnet-18), the acquisition batch severely degenerates. In contrast \texttt{GauSS} does not query based on the representation exclusively. Therefore switching experiment components does not affect the overall protocol performance. 
Finally, we discuss the behavior of \texttt{GauSS} in settings with high class imbalances on CURE-TSR. In our experiments, \texttt{GauSS} outperforms on two of the three architecture choices and consistently shows positive improvements over the random baseline (i.e. positive integral values). Similar to our previous observations, competitive strategies are strong in one setting but underperform the random baseline in another. For instance, entropy sampling underperforms the random baseline on resnet-18 but outperforms on resnet-34. We follow that \texttt{GauSS} proposes consistent and accurate data samples even when high imbalances are present.

In addition to Table~\ref{table:results-overall}, we provide absolute values of the different strategies (as well as the random baseline) in Table~\ref{table:results-absolute}. Specifically, we show accuracy values in rounds one, four, eight, and nine on CIFAR10, as well as its corrupted variant CIFAR10-C. We choose this constellation as it represents one early round, one intermediate round, and two late-stage rounds. Further, we conduct a t-test to determine the statistical significance of our results, where one refers to a statistically significant difference to \texttt{GauSS}, and zero to a statistically insignificant difference with \texttt{GauSS}.  We note, that \texttt{GauSS} matches or outperforms the competing strategies across all rounds: \texttt{GauSS} either outperforms or is statistically insignificant with respect to the outperforming strategy. In contrast, existing strategies are either competitive in early rounds or later rounds but rarely show consistent performance across the entire active learning experiment.
%


\begin{table*}[!ht]
	\centering
	\caption{Area under difference curve with reference to random sampling over several datasets, query strategies, and architectures. Positive, and higher numbers are better. Negative numbers imply that the approach has a lower accuracy curve than the random baseline. The spread represents the standard deviation. Random reference area shows the total area under the random baseline accuracy curve.}
	\label{table:results-overall}
	\begin{tabular}{ |p{1.9cm}|p{1.9cm}||p{1.6cm}|p{1.6cm}|p{1.6cm}|p{1.6cm}|p{1.8cm}|p{1.6cm}|}
		\hline
		\multicolumn{2}{|c||}{}   & \multicolumn{3}{|c|}{Out-Of-Distr.} & \multicolumn{3}{|c|}{In-Distr.} \\
		\hline
		
		\multicolumn{1}{|c|}{Architectures} & \multicolumn{1}{|c||}{Algorithms} & \multicolumn{1}{|c|}{CIFAR10-C}  & \multicolumn{1}{|c|}{CINIC10}  & \multicolumn{1}{|c|}{STL10} & \multicolumn{1}{|c|}{CIFAR10} & \multicolumn{1}{|c|}{CINIC10} & \multicolumn{1}{|c|}{CURE-TSR}\\
		\hline
        \multirow{6}{*}{ResNet-18}  
                                    & Rand. Reference           & 649.53     & 666.06      & 399.29     & 831.19       & 636.35     & 972.48\\
                                    & ALBL \cite{hsu2015active}       & 10.19 ± 8.24     & 5.66 ± 8.08      & -4.55 ± 4.66     & 7.25 ± 6.86       & -2.54 ± 5.12     & 22.80 ± 1.48\\
                                    & Coreset \cite{sener2017active}  & -41.87 ± 2.02    & -35.07 ± 2.13    & -20.58 ± 3.14    & -41.04 ± 0.71    & -48.54 ± 3.11     & -38.60 ± 0.83            \\
                                    & Entropy \cite{wang2014new}      & 7.83 ± 1.03      & 8.58 ± 0.76      & 2.16 ± 3.79      & 0.45 ± 0.34      & -6.09 ± 3.44      & -25.02 ± 8.55            \\
                                    & L. Conf. \cite{wang2014new}     & 16.23 ± 1.17     & 13.47 ± 0.12     & 3.20 ± 1.42      & 9.93 ± 0.94      & -2.54 ± 1.22      & 5.74  ± 6.28              \\
                                    & BatchBald \cite{batchbald}      & -20.98 ± 2.40    & -19.93 ± 0.75    & -14.71 ± 0.01    & -17.49 ± 2.33    & -356.14 ± 5.30   & -2.56 ± 3.17                \\
                                    & \bf{GauSS}                & $\mathbf{17.76\pm1.61}$ & $\mathbf{13.72\pm0.13}$ & $\mathbf{7.17\pm4.48}$ & $\mathbf{17.77\pm0.89}$ & $\mathbf{5.23\pm0.47}$ &                                        $\mathbf{33.55\pm6.13}$ \\
                                                                        		\hline
        \multirow{6}{*}{DenseNet-121}
                                     & Rand. Reference           & 653.55     & 674.07      & 413.96     & 860.81       & 661.66     & 1018.35\\
                                     & ALBL \cite{hsu2015active}       & 7.69 ± 1.41      & 6.67 ± 2.45      & -8.56 ± 2.26     & 3.28 ± 3.08      & -4.58 ± 2.61     & 30.91 ± 9.30             \\
                                     & Coreset \cite{sener2017active}  & -23.40 ± 2.32    & -14.36 ± 0.70    & -24.98 ± 1.94    & -21.74 ± 2.89    & -12.72 ± 0.77    & 31.66 ± 6.83             \\
                                     & Entropy \cite{wang2014new}      & 1.77 ± 0.68      & 2.52 ± 2.75      & -10.25 ± 5.34    & -7.76 ± 4.20     & -23.17 ± 3.92    & 8.33 ± 6.34             \\
                                     & L. Conf. \cite{wang2014new}     & -0.20 ± 1.59     & 3.81 ± 1.89      & -11.48 ± 4.65    & -3.57 ± 4.17     & -17.19 ± 5.08    & 43.25 ± 8.85             \\
                                     & BatchBald \cite{batchbald}      & -8.85 ± 0.47     & -11.98 ± 0.01    & -1.15 ± 1.79     & -10.51 ± 0.64    & -383.24 ± 19.51  & $\mathbf{60.13\pm9.30}$               \\
                                     & \bf{GauSS}                      & $\mathbf{13.64\pm1.96}$ & $\mathbf{10.01\pm0.52}$ & $\mathbf{-1.28\pm0.64}$ & $\mathbf{11.46\pm0.18}$ & $\mathbf{4.52\pm2.61}$  &                                 53.25 ± 5.47 \\
                                                                        		\hline
        \multirow{6}{*}{ResNet-34}   
                                     & Rand. Reference           & 705.04     & 708.21      & 427.31     & 893.57       & 687.10     & 1018.21\\
                                     & ALBL \cite{hsu2015active}       & 7.56 ± 0.09             & 9.67 ± 2.42     & 0.41 ± 4.06           & 13.96 ± 2.40     & -4.55 ± 3.78            & 28.63 ± 4.16             \\
                                     & Coreset \cite{sener2017active}  & -8.25 ± 0.17            & -7.18 ± 0.24    & -9.23 ± 4.28          & -5.21 ± 0.27     & -14.16 ± 3.70     & -9.40 ± 6.88             \\
                                     & Entropy \cite{wang2014new}      & 14.80 ± 0.16            & 15.31 ± 0.53    & $\mathbf{5.93\pm1.34}$& 12.51 ± 1.03     & -12.49 ± 3.61           & 10.01 ± 6.47             \\
                                     & L. Conf. \cite{wang2014new}     & 17.52 ± 1.21            & 16.60 ± 1.33    & 4.91 ± 5.17           & 15.99 ± 0.36     & -4.40 ± 0.83            & $\mathbf{47.01\pm3.83}$  \\
                                     & BatchBald \cite{batchbald}      & -25.90 ± 0.39           & -25.84 ± 0.93   & -11.85 ± 0.76         & -26.10 ± 1.49    & -391.38 ± 11.78  & 0.93 ± 3.13               \\
                                     & \bf{GauSS}                      & $\mathbf{18.14\pm0.96}$ & $\mathbf{16.58\pm1.17}$ & 2.12 ± 2.66        & $\mathbf{19.30\pm1.16}$ & $\mathbf{9.34\pm1.04}$ &                                 25.44 ± 9.62             \\
		\hline
	\end{tabular}
\end{table*}

\begin{table*}[!ht]
	\centering
	\caption{Accuracy values and binary statistical significance over active learning rounds one, four, eight, and nine on CIFAR10 and CIFAR10-C. A significance of one indicates a statistically significant difference to GauSS while zero indicates a statistically insignificant difference with \texttt{GauSS}. We show our results on ResNet-18 and DenseNet-121.}
	\label{table:results-absolute}
	\begin{tabular}{ |p{0.2cm}|p{1.9cm}|p{1.9cm}||p{1.0cm}|p{0.5cm}|p{1.0cm}|p{0.5cm}|p{1.0cm}|p{0.5cm}|p{1.0cm}|p{0.5cm}|}
		\hline
		\multicolumn{3}{|c||}{}   & \multicolumn{2}{|c|}{1152 Samples} & \multicolumn{2}{|c|}{4224 Samples} & \multicolumn{2}{|c|}{8320 Samples} & \multicolumn{2}{|c|}{9344 Samples} \\
		\hline
		
		 \multicolumn{1}{|c|}{} & \multicolumn{1}{|c|}{Dataset} & \multicolumn{1}{|c||}{Algorithms} & \multicolumn{1}{|c|}{Acc.}  & \multicolumn{1}{|c|}{Stat. Sign.}  & \multicolumn{1}{|c|}{Acc.}  & \multicolumn{1}{|c|}{Stat. Sign.}  & \multicolumn{1}{|c|}{Acc.}  & \multicolumn{1}{|c|}{Stat. Sign.}  & \multicolumn{1}{|c|}{Acc.}  & \multicolumn{1}{|c|}{Stat. Sign.} \\
		\hline
  \parbox[t]{2mm}{\multirow{6}{*}{\rotatebox[origin=c]{90}{ResNet-18}}}  	& \multirow{6}{*}{CIFAR10}  
                                    & Random                          & 42.48 & 0   & 57.05 & 0   & 63.89 & 1   & 65.02 & 1   \\
                                    & & ALBL \cite{hsu2015active}       & 39.14 & 1   & 57.16 & 1   & 64.39 & 0   & 65.89 & 0   \\
                                    & & Coreset \cite{sener2017active}  & 38.00 & 1   & 52.07 & 1   & 61.96 & 1   & 63.21 & 1   \\
                                    & & Entropy \cite{wang2014new}      & 37.35 & 1   & 54.90 & 1   & 64.02 & 1   & 65.00 & 1   \\
                                    & & L. Conf. \cite{wang2014new}     & 38.90 & 1   & 56.11 & 1   & 64.29 & 0   & 65.47 & 0   \\
                                    & & BatchBald \cite{batchbald}      & \bf{43.36}  & 0  & \bf{57.25} & 0   & 63.20 & 1   & 64.10 & 1   \\
                                    & & \bf{GauSS}                      & 42.00 & Baseline & 57.06 & Baseline & \bf{64.76} & Baseline & \bf{66.11} & Baseline \\
		\hline
   \parbox[t]{2mm}{\multirow{6}{*}{\rotatebox[origin=c]{90}{ResNet-18}}}  	& \multirow{6}{*}{CIFAR10-C}  
                                    & Random                          & \bf{33.27} & 0   & 44.19 & 0   & 49.49 & 1   & 50.72 & 1\\
                                    & & ALBL \cite{hsu2015active}       & 31.01 & 1   & \bf{44.64} & 0   & \bf{50.46} & 0   & 51.60 & 0\\
                                    & & Coreset \cite{sener2017active}  & 29.83 & 1   & 39.49 & 1   & 47.44 & 1   & 48.44 & 1\\
                                    & & Entropy \cite{wang2014new}      & 30.17 & 1   & 42.88 & 1   & 50.28 & 0   & 51.33 & 0\\
                                    & & L. Conf. \cite{wang2014new}     & 31.15 & 1   & 43.68 & 1   & 50.45 & 0   & 51.59 & 0\\
                                    & & BatchBald \cite{batchbald}      & 33.97 & 0   & 43.69 & 0   & 48.83 & 1   & 49.92 & 1\\
                                    & & \bf{GauSS}                      & 33.09 & Baseline & 44.43 & Baseline & 50.43 & Baseline & \bf{51.72} & Baseline\\
		\hline
  \parbox[t]{2mm}{\multirow{6}{*}{\rotatebox[origin=c]{90}{DenseNet-121}}}  	&
        \multirow{6}{*}{CIFAR10}  
                                    & Random                          & \bf{44.17} & 0   & 59.27 & 0   & 66.73 & 0   & 67.63 & 1   \\
                                    & & ALBL \cite{hsu2015active}       & 38.53 & 1   & 59.34 & 0   & \bf{67.22} & 0   & 68.81 & 0   \\
                                    & & Coreset \cite{sener2017active}  & 36.13 & 1   & 58.08 & 1   & 65.89 & 1   & 67.30 & 1   \\
                                    & & Entropy \cite{wang2014new}      & 37.53 & 1   & 57.54 & 1   & 66.94 & 0   & 68.17 & 0   \\
                                    & & L. Conf. \cite{wang2014new}     & 39.06 & 1   & 58.07 & 1   & 66.62 & 0   & 67.62 & 1   \\
                                    & & BatchBald \cite{batchbald}      & 43.84 & 0   & \bf{60.45} & 0   & 66.17 & 0   & 66.84 & 1   \\
                                    & & \bf{GauSS}                      & 43.38 & Baseline & 59.53 & Baseline & 67.02 & Baseline & \bf{69.01} & Baseline \\
		\hline
        \parbox[t]{2mm}{\multirow{6}{*}{\rotatebox[origin=c]{90}{DenseNet-121}}}  	& \multirow{6}{*}{CIFAR10-C}  
                                    & Random                          & \bf{33.21} & 0   & 44.74 & 0   & 50.31 & 1   & 51.42 & 1   \\
                                    & & ALBL \cite{hsu2015active}       & 29.53 & 1   & \bf{45.27} & 0   & 51.26 & 0   & 52.64 & 0   \\
                                    & & Coreset \cite{sener2017active}  & 26.94 & 1   & 43.25 & 1   & 49.52 & 1   & 50.55 & 1   \\
                                    & & Entropy \cite{wang2014new}      & 29.29 & 1   & 43.69 & 1   & 50.71 & 1   & 51.72 & 1   \\
                                    & & L. Conf. \cite{wang2014new}     & 30.07 & 1   & 44.18 & 0   & 50.19 & 1   & 51.63 & 1   \\
                                    & & BatchBald \cite{batchbald}      & 32.87 & 0   & 45.43 & 0   & 50.12 & 1   & 50.82 & 0   \\
                                    & & \bf{GauSS}                      & 33.06 & Baseline & 44.75 & Baseline & \bf{51.65} & Baseline & \bf{52.76} & Baseline \\
		\hline
	\end{tabular}
\end{table*}

\begin{figure*}[!h]
    \begin{center}
        \includegraphics[scale=0.38]{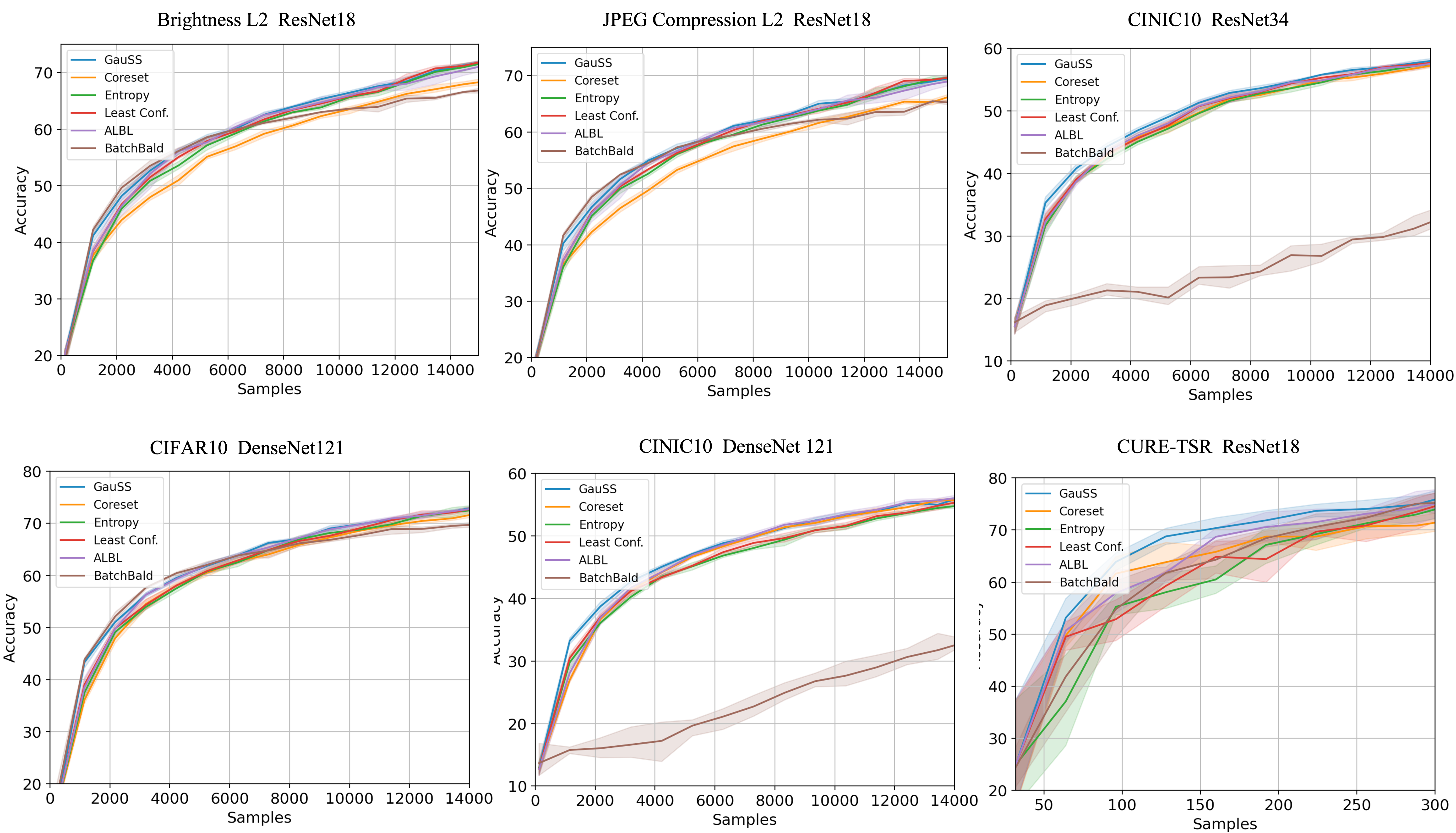}
    \end{center}
    \caption{Learning curves of in-distribution, as well as out-of-distribution experiments with the resnet-18, resnet-34, and densenet-121. Top: Out-of-distribution experiments with resnet-18 on brightness level two (CIFAR10-C), JPEG compression level five (CIFAR10-C), and in-distribution experiment on CINIC10 with resnet-34. Bottom: In-distribution experiments with CIFAR10 with desnsenet-121, CINIC10 with densenet-121, and CURE-TSR with resnet-18.}
    \label{fig:lcs-final}
\end{figure*}
\subsection{Learning Curves}
We further plot the learning curves of several experiment configurations in Figure~\ref{fig:lcs-final}. Specifically, we show the results of several configurations using a resnet-18 architecture on in-distribution, as well as out-of-distribution experiments. As out-of-distribution examples, we plot the CIFAR10-C corruptions brightness level two, JPEG compression level five, as well as the CINIC10 test set when training on CIFAR10. Further, we plot the in-distribution datasets CIFAR10, CININC10, and CURE-TSR. The learning curves, further confirm our observations from Table~\ref{table:results-overall}. \texttt{GauSS} qualitatively outperforms the other strategies over a large variety of in-distribution, and out-of-distribution experiments. We further note, that the performance of \texttt{GauSS} is especially strong in early rounds where the training set contains few samples. We reason that the representation is especially inaccurate for early active learning rounds and that resulting importance metric is severly affected for the competitive strategies. For later rounds, the model develops a more accurate representation space and we observe \texttt{GauSS} outperforming by a significantly smaller margin. A further explanation for the superior performance is the targeted sampling approach in \texttt{GauSS}. As previously discussed, representation noise can be beneficial in scenarios where accurate importance definitions result in difficult outlier selections that can potentially damage the model representation (see our experiments with optimal representations in Section~\ref{sec:optimal-strat-lcs}). This is especially true for the two protocols least confidence sampling, and entropy sampling that have a forcefully added random query noise due to the limited representation capabilities of the model. In contrast, \texttt{GauSS} adds targeted (or biased) noise that, in combination with accurate importance measurements, results in a more conclusive data representation when the representation is severely susceptible to outlier samples. 

\begin{figure*}[!th]
    \begin{center}
        \includegraphics[scale=0.38]{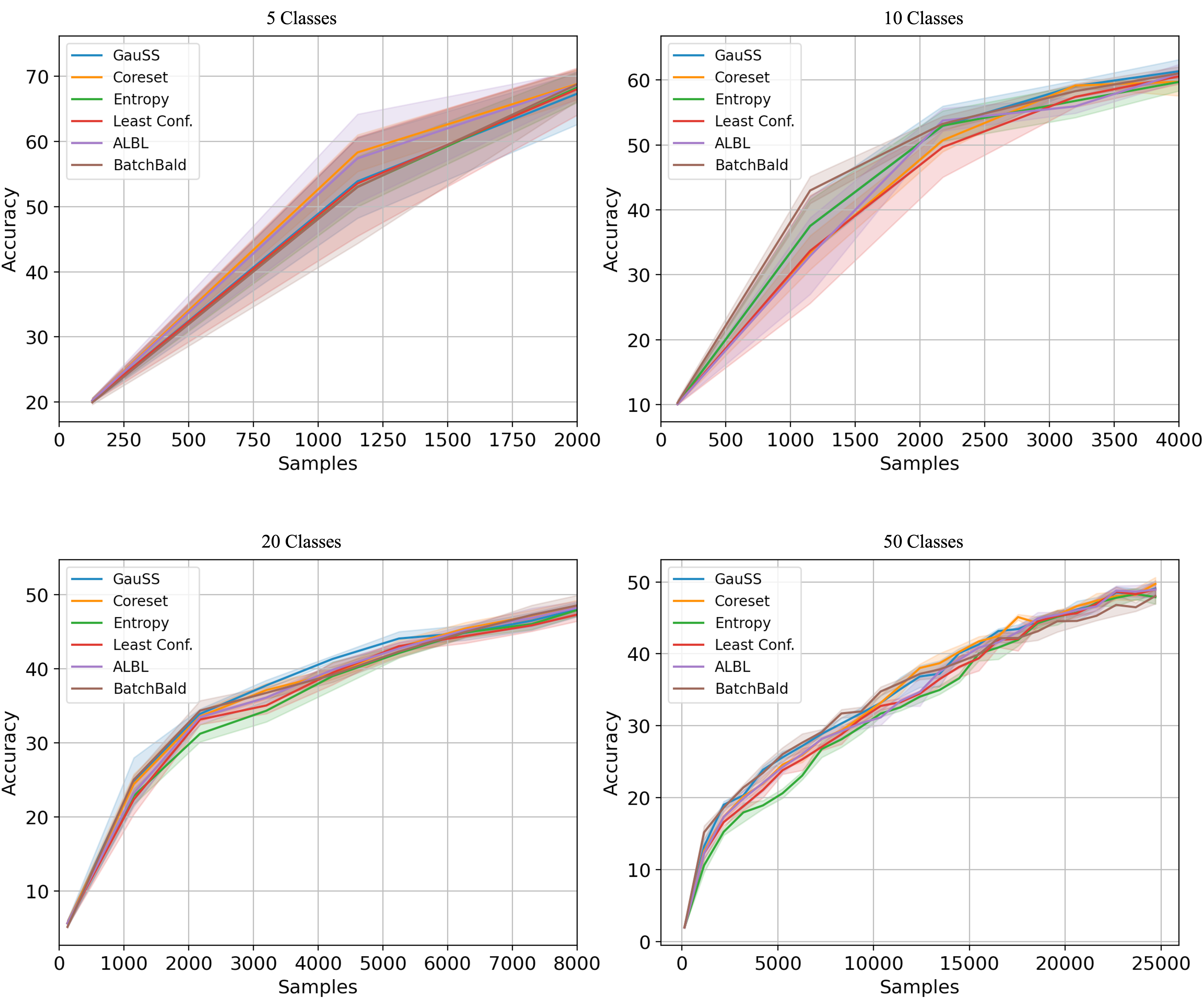}
    \end{center}
    \caption{Incremental analysis on CIFAR100. We perform active learning with a resnet-18 architecture on 5, 10, 20, and 50 of the 100 CIFAR100 classes respecitvely. We compare \texttt{GauSS} with coreset, entropy, least confidence, active-learning-by-learning, and batchbald.}
    \label{fig:lcs-incremental}
\end{figure*}
While we only show a small subset of learning curves, we note that this behavior is consistent across the other experiment constellations as well but are omitted due to space limitations. With 16 different CIFAR10-C corruptions at two levels as well as the remaining benchmarks this amounts to 555 different active learning curves. In summary, we can report an improvement of up to 5\% compared to the other strategies.

\subsection{Incremental Analysis}
Within this subsection, we investigate the performance of \texttt{GauSS} in settings with higher task complexity. For this purpose, we consider an incremental analysis where we perform active learning on a subset of the CIFAR100 dataset with 5, 10, 20, and 50 of the 100 classes respectively. In our experiments, we use a resnet-18 architecture and opt for a similar setup as our previous experiments on CIFAR10. In particular, we start with an initial training set size of 128 and query 1024 samples each round until the entire unlabeled pool is fully annotated. We show the resulting learning curves of the four data subsets in Figure~\ref{fig:lcs-incremental}. We note that settings with less classes contain fewer samples in the data pool and are therefore fully queried in less rounds. 

Overall, \texttt{GauSS} shows favorable qualities with a higher number of tasks. While lower task quantities do not show distinguishable trends, \texttt{GauSS} performs well in complex settings with higher amounts of tasks. In particular, we note that \texttt{GauSS} performs well regardless of the active learning round. In contrast, the competitive strategies show a strong performance in early or later active learning rounds exclusively. As an example, entropy and least confidence sampling perform well in later rounds and underperform in early stages. On the other hand, Batchbald is particularly strong in early stages but is outperformed later stages where the data representations are more mature.

%% file: sections/Conclusion.tex
\section{Conclusion}
\label{sec:conclusion}
In this paper, we introduced a grounded definition of information content for active learning protocols. In contrast to existing approaches, our definition is not based on the representation directly but defines importance with decision boundary shifts - or neural network ``forgetting". For practical usage, we approximate ``forgetting" with prediction switches and find that prediction switches produce accurate importance scores when compared to other definitions. Finally, we develop a novel acquisition function that samples interactively with a gaussian mixture model. We validate our algorithm empirically with exhaustive experiments and compare against popular protocols used in practical active learning pipelines. Overall, \texttt{GauSS} is robust to setup changes, performs favorably in out-of-distribution settings, and achieves up to 5\% improvement in terms of accuracy.